\documentclass[10pt]{article}
\usepackage{radio-report}
\usepackage{radio-examples}
\usepackage{pifont,xspace,float}
\usepackage{needspace}

\usepackage{bm}

\def\eqref#1{equation~\ref{#1}}

\def\1{\bm{1}}

\DeclareMathAlphabet{\mathsfit}{\encodingdefault}{\sfdefault}{m}{sl}
\SetMathAlphabet{\mathsfit}{bold}{\encodingdefault}{\sfdefault}{bx}{n}

\newcommand{\ms}[2]{\ensuremath{#1_{\,\pm #2}}}
\definecolor{greyC}{RGB}{180,180,180}
\definecolor{greyL}{RGB}{235,235,235}
\definecolor{Gray}{gray}{0.9}
\definecolor{mydarkred}{rgb}{0.6,0,0}
\definecolor{myblue}{HTML}{268BD2}
\definecolor{mygreen}{HTML}{658354}
\definecolor{orangeinplot}{HTML}{e29c7a}
\definecolor{purpleinplot}{HTML}{7676a4}
\definecolor{greeninplot}{HTML}{288308}
\definecolor{mydarkblue}{rgb}{0,0.08,0.45}
\definecolor{hypogainsboro}{HTML}{F1F3F5}

\providecommand{\method}{\textsc{ToolCompass}\xspace}
\newcommand{\promptcaption}[2]{\begingroup\captionsetup{type=figure}\captionof{figure}{#1}\label{#2}\endgroup}

\AtBeginEnvironment{thebibliography}{\interlinepenalty=10000}

\renewcommand{\reporttitle}{ToolCompass: Guiding Tool Trialing, Not Suppressing It}
\renewcommand{\reportshorttitle}{ToolCompass: Guiding Tool Trialing, Not Suppressing It}
\renewcommand{\reportdate}{September 2026}
\reportauthor[1]{Junlin Fang}
\reportauthor{Chong Zhang}
\reportauthor[1]{Do Nguyen-Thanh}
\reportauthor[2]{Xiaogang Xu}
\reportauthor[3]{Zhen Fang}
\reportauthor[1,*]{Sean Du}
\reportaffil[1]{College of Computing and Data Science, Nanyang Technological University}
\reportaffil[2]{Zhejiang University}
\reportaffil[3]{Australian Artificial Intelligence Institute, University of Technology Sydney}
\reportaffil[*]{Corresponding author}
\renewcommand{\reportcontact}{\href{mailto:junlin001@e.ntu.edu.sg}{\nolinkurl{junlin001@e.ntu.edu.sg}};\enspace\href{mailto:xuefeng.du@ntu.edu.sg}{\nolinkurl{xuefeng.du@ntu.edu.sg}}}
\renewcommand{\reportabstract}{Large language model (LLM) agents must generalize from tools seen during training to unseen tools at deployment. A key challenge is \emph{tool trialing}, i.e., excessive trials waste the interaction budget, whereas selective trials enable exploration of unfamiliar tools. Existing outcome-based post-training leaves wasteful trials unguided, while turn-level supervision may suppress necessary exploration. We introduce \method, a post-training framework that guides tool trialing by organizing tool-call representations according to shared functions. Specifically, \method models each function class as a von Mises--Fisher distribution and jointly reduces intra-function variation across domains and increases inter-function separation. This structure transfers experience from seen tools to functionally similar unseen tools, directing exploration away from unrelated alternatives. \method requires no ground-truth call traces or unseen-tool access and incurs no inference overhead. Experiments on AppWorld and FTRL show consistent gains across GRPO, RFT, and DMPO. \method improves AppWorld OOD task success by up to 10.71 percentage points over vanilla post-training and performs best among competitive baselines on both benchmarks.
}
\hypersetup{pdftitle={ToolCompass: Guiding Tool Trialing, Not Suppressing It},pdfauthor={Junlin Fang, Chong Zhang, Do Nguyen-Thanh, Xiaogang Xu, Zhen Fang, Sean Du}}
\begin{document}
\makereporttitle
\section{Introduction}
\label{sec:intro}

\begin{figure}[t]
    
    \centering
    \includegraphics[width=\linewidth,trim=0 0 0 0,clip]{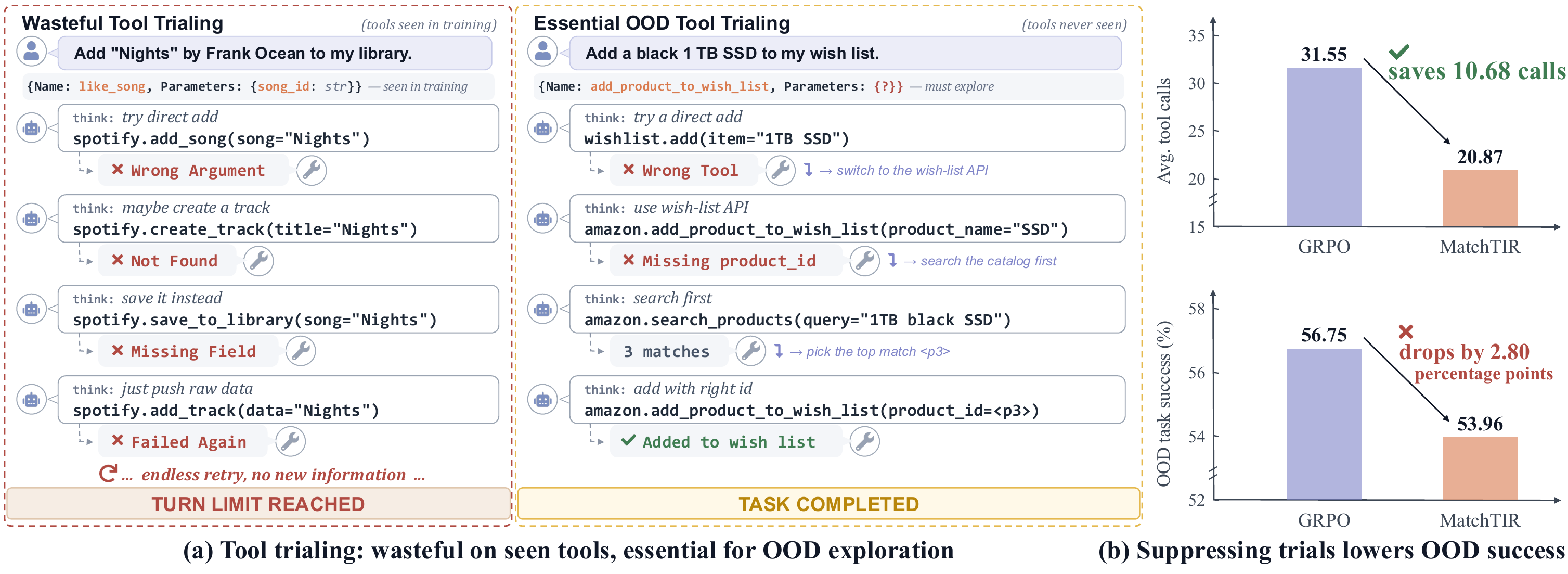}
    
    \caption{\textbf{Tool trialing should be guided rather than suppressed.} (a) Excessive trialing wastes the interaction budget to find the needed tool ``like\_song", whereas selective trialing enables exploration of unfamiliar tools of ``add\_product\_to\_wish\_list". (b) On held-out AppWorld applications, MatchTIR makes fewer calls than GRPO but achieves lower task success, revealing the cost of suppressed exploration. 
    }
    \label{fig:motivation}
    
\end{figure}

 Large language model (LLM) agents interact with external environments through tools such as search engines and application APIs~\citep{yao2023react,schick2023toolformer,patil2024gorilla,qin2024toolllm,qu2025tool}. A reliable agent should not only use tools seen during training, but also transfer learned functions to unseen tools in out-of-distribution (OOD) environments~\citep{tang2023toolalpaca,mekala2024toolverifier,he2025gentool}. For example, experience with \texttt{amazon.search\_products} should help the agent invoke \texttt{spotify.search\_songs}. In multi-turn environments, an agent may try a plausible call, observe an error, and revise its tool choice or arguments. We refer to this trial-and-correction behavior as \emph{tool trialing}~\citep{gao2026teaching}. Unnecessary trials consume the interaction budget and can cause solvable tasks to fail~\citep{liu2025budget,mansoor2026verified}. Yet trialing might be essential for unfamiliar tools, whose behavior must be learned through environment feedback. The agent must therefore reduce wasteful calls without losing the exploration required for OOD transfer (Figure~\ref{fig:motivation}a).

Existing post-training methods do not effectively resolve this trade-off. Outcome-based methods for tool use~\citep{qian2026toolrl,jin2025searchr1,feng2026retool}, such as GRPO~\citep{shao2024deepseekmath}, assign the same trajectory-level advantage across tool turns, learning wasteful trials together with useful solution steps. Recent methods instead introduce turn-level supervision~\citep{yu2025steptool,zeng2025reinforcing}, either by matching predicted calls to a ground-truth trace (MatchTIR~\citep{qu2026matchtir}) or estimating each turn's contribution to the final outcome (TRACE~\citep{tao2026trace}; StepTool~\citep{yu2025steptool}). Despite the promise, such supervision can discourage deviations from known trajectories, including exploration that becomes necessary for unseen tools. Our pilot study confirms this limitation on held-out AppWorld applications~\citep{trivedi2024appworld}: MatchTIR substantially reduces tool calls and rarely reaches the turn limit, yet achieves lower task success than GRPO (Figure~\ref{fig:motivation}b). This motivates the central question of our work: 

\begin{center}
\textbf{How can post-training enable OOD  exploration without reinforcing wasteful tool trialing?}

\end{center}

To understand this failure, we examine the rollouts and tool-call representations of a Qwen3.5-9B~\citep{qwen35blog} agent post-trained with GRPO on AppWorld. When an unseen tool shares a function with a seen tool, the agent does not exploit this similarity to narrow its exploration. Instead, it trials unrelated tools before reaching a useful call (related examples in Appendix~\ref{app:qualitative}). The representation space exhibits a corresponding pattern that \textit{calls from the same domain lie close together despite performing different functions, whereas calls sharing a function remain far apart across domains} (Figure~\ref{fig:tease_tsne}a). The learned representations therefore encode domain identity rather than the function-level commonality needed to transfer experience across tools.

Motivated by this finding, we propose \method, a vMF-based post-training framework that shapes tool-call representations by shared function. Our key idea is to promote low intra-function variation across domains and high inter-function separation. Specifically, \method maps each tool-call representation onto the unit hypersphere.  Each function class can then be naturally modeled as a von Mises--Fisher (vMF) distribution~\citep{mardia2009directional} centered at a prototype (Figure~\ref{fig:framework}). A variation loss draws calls toward their function prototype, while a separation loss keeps different prototypes apart. The resulting objective is optimized jointly with the original post-training objective. By organizing the policy's hidden representations around tool functions, \method makes experience with seen tools more transferable to functionally similar unseen tools, directing exploration away from unrelated alternatives (Figure~\ref{fig:tease_tsne}b).

Importantly, \method requires only the function-class assignments of seen tools, without ground-truth call traces, access to unseen tools, or frozen turn-level models during training. It can be readily combined with GRPO~\citep{shao2024deepseekmath}, RFT~\citep{yuan2023rft}, and DMPO~\citep{shi2024dmpo}. The projection head and prototypes are removed after post-training, introducing no additional inference overhead. \method therefore guides \emph{where} the agent trials rather than directly penalizing \emph{how much} it trials. Extensive experiments on AppWorld and FTRL~\citep{ye2026feedback}, using Qwen3.5-4B and Qwen3.5-9B, show that \method consistently improves task success across all three post-training methods, with strong performance on OOD tasks requiring unseen tools. On AppWorld, \method improves OOD task success over vanilla GRPO by up to 8.87 percentage points, reaching 70.02\% with Qwen3.5-9B. Further analyses show that \method produces shorter trajectories, reduces trials of unrelated tools, increases the adoption of task-relevant unseen tools, and organizes representations by shared function rather than domain (Section~\ref{sec:exp:analysis}). Our key contributions are summarized as follows:

\begin{itemize}
\item We identify a tool-trialing trade-off in OOD tool use, i.e.,  outcome rewards leave wasteful trials unguided, whereas strict turn-level supervision can suppress necessary exploration. We further connect this failure to representations organized by domain rather than function.
\item We propose \method, a vMF-based framework that reduces intra-function variation and increases inter-function separation during post-training, guiding exploration without ground-truth call traces, unseen-tool access, or additional inference computation.
\item Extensive experiments on AppWorld and FTRL demonstrate consistent improvements across models and post-training objectives, including on OOD tasks, alongside more efficient trialing and stronger adoption of unseen tools.
\end{itemize}

\section{Problem Setup}
\label{sec:setup}

Formally, we describe the tool-use agent, post-training and deployment settings, and learning goal.
 
\textbf{Post-training for tool-use agents.}
A tool-use agent is a policy $\pi_\theta$ that interacts with an environment over a trajectory $\tau=(s_1,a_1,o_1,\ldots,s_T,a_T,o_T)$. At turn $t$, the state $s_t$ contains the user request, interaction history, and specifications of the available tools; the action $a_t$ is either a language response or a call to tool $u_t\in\mathcal{U}$; and the observation $o_t$ is the resulting environment feedback, such as an execution result or error message. 

Tools from different domains may implement the same function. For example, \texttt{amazon.search\_products} and \texttt{spotify.search\_songs} both perform \textsc{Search}. We group such tools into \textit{function classes} and denote the function class of tool $u$ by $c(u)\in\mathcal{C}$. The agent is post-trained on tasks involving a fixed set of seen tools $\mathcal{U}_{\mathrm{train}}$ using an objective $\mathcal{L}_{\mathrm{post}}(\theta)$.

\textbf{OOD tool generalization.}
At deployment, the agent encounters in-distribution (ID) tasks solvable with tools in $\mathcal{U}_{\mathrm{train}}$, as well as OOD tasks requiring tools from an unseen set $\mathcal{U}_{\mathrm{unseen}}$, where $\mathcal{U}_{\mathrm{train}}\cap\mathcal{U}_{\mathrm{unseen}}=\varnothing$. These unseen tools belong to held-out domains but may share similar functions with the seen tools. Their specifications are provided to the agent only at deployment through $s_t$; \textit{neither the tools nor their annotations are available} during post-training. Our goal is to transfer function-level knowledge from $\mathcal{U}_{\mathrm{train}}$ to $\mathcal{U}_{\mathrm{unseen}}$ while preserving performance on ID tasks.

\Needspace{20\baselineskip}
\begin{wrapfigure}{r}{0.57\textwidth}
    \centering
    
    \includegraphics[width=\linewidth]{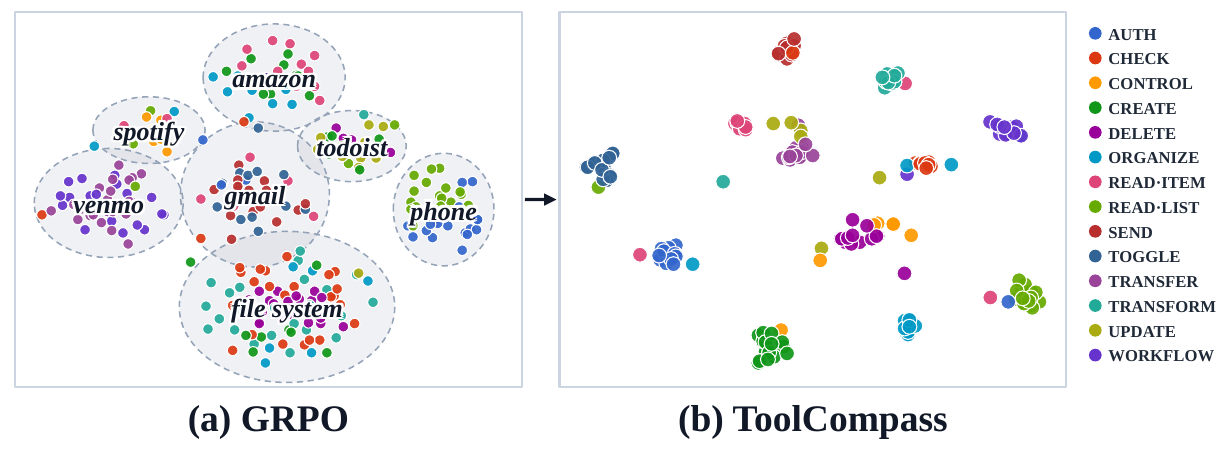}
    
    \caption{  \textbf{Tool-call representations on AppWorld.} After GRPO post-training (a), representations cluster primarily by application domain (dashed ellipses) rather than function class (colors). In contrast, \method (b) aligns tool calls sharing the same function across domains. }
    \label{fig:tease_tsne}
    
\end{wrapfigure}
\textbf{Challenge and learning goal.} A tool-use trajectory may contain useful trials of functionally plausible tools as well as unproductive trials of unrelated tools. The latter consume the interaction budget and can cause a solvable task to fail~\citep{liu2025budget,mansoor2026verified}, whereas the former provide feedback needed to use unfamiliar tools. This challenge is reflected in the representation space learned by standard post-training: as shown in Figure~\ref{fig:tease_tsne}(a), GRPO organizes tool calls primarily by domain (e.g., different applications) rather than by shared function, providing little structure for transferring experience to unseen tools. Penalties that indiscriminately discourage trialing may instead suppress both useful and unproductive calls. We therefore aim to learn function-level representations, as shown in Figure~\ref{fig:tease_tsne}(b), that guide trialing toward functionally relevant tools without suppressing the exploration required for OOD generalization.

\FloatBarrier
\section{Method}
\label{sec:method}
\begin{figure}[t]
    
    \centering
    \includegraphics[width=\linewidth,trim=0 0 0 0,clip]{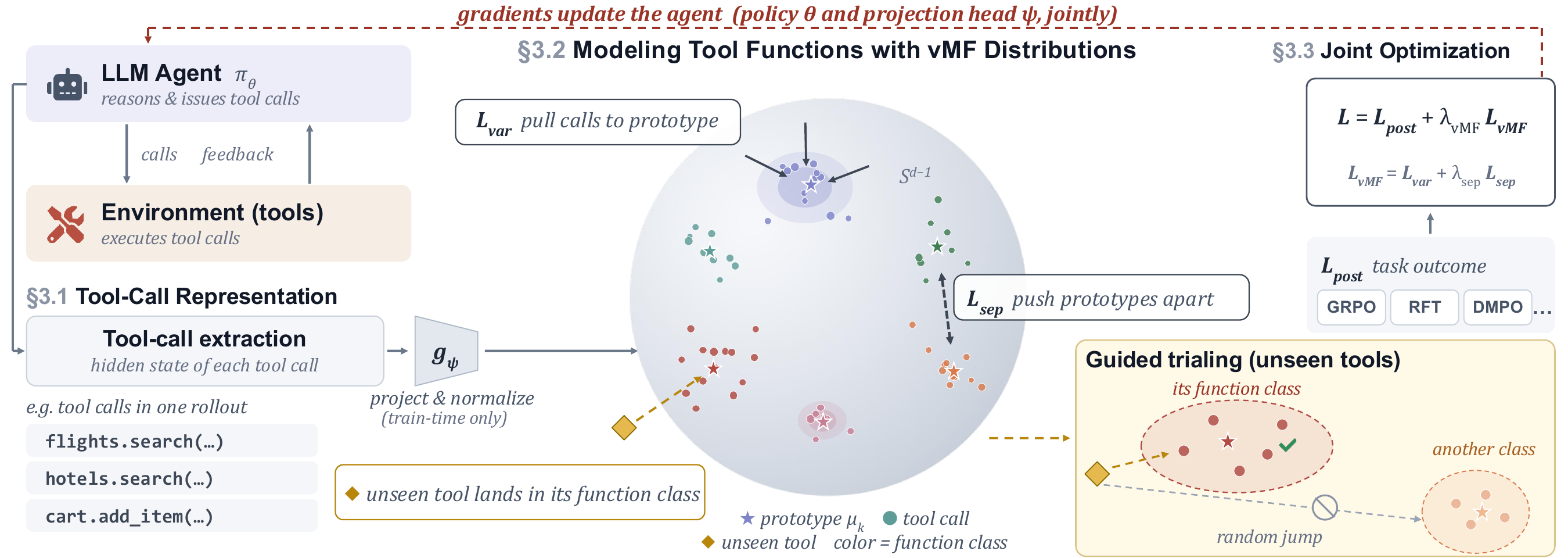}
    
    \caption{ \textbf{Overview of \method.} For each tool call, \method extracts its final-token representation and maps it onto the unit hypersphere. Calls implementing the same function are aligned with a shared vMF prototype, while calls implementing different functions are separated. The resulting objective is optimized jointly with the post-training objective. The projection head and prototypes are not used at deployment.}
    \label{fig:framework}
    
\end{figure}

    \textbf{Framework overview.}
Our framework \method guides tool trialing by shaping the representations of tool calls according to their functions. Central to our framework is to reduce variation among calls implementing the same function across domains, while separating calls implementing different functions. As illustrated in Figure~\ref{fig:framework}, \method first extracts and normalizes tool-call representations (Section~\ref{sec:method:repr}). It then models each function class with a von Mises--Fisher (vMF) distribution and optimizes intra-function variation and inter-function separation (Section~\ref{sec:method:vmf}). The resulting objective is trained jointly with the original post-training objective (Section~\ref{sec:method:joint}).

\subsection{Tool-Call Representation}
\label{sec:method:repr}

\textbf{Representation extraction.}
For each tool call $a_t$ in a sampled trajectory, we extract the  hidden representation from the agent (with parameter $\theta$):
\begin{equation}
\mathbf{h}_t=H^{(l)}_\theta(s_t,a_t)\in\mathbb{R}^{d_h},
\label{eq:hidden}
\end{equation}
where $H^{(l)}_\theta$ returns the hidden state at layer $l$ and the final token of $a_t$, and $d_h$ is the hidden dimension. Since the final token attends to the preceding context, $\mathbf{h}_t$ summarizes the selected tool, its arguments, and the current interaction state. We study the choice of layer $l$ in Section~\ref{sec:experiments}.

\textbf{Hyperspherical normalization.}
A lightweight projection head $g_\psi:\mathbb{R}^{d_h}\rightarrow\mathbb{R}^{d}$ maps $\mathbf{h}_t$ to
$
\mathbf{z}_t=
\frac{g_\psi(\mathbf{h}_t)}
{\lVert g_\psi(\mathbf{h}_t)\rVert_2}
\in\mathbb{S}^{d-1}.
$
The unit norm places all tool-call representations on the unit hypersphere, where angular similarity characterizes their relationships and the vMF distribution is naturally defined~\citep{mardia2009directional}.

\subsection{Modeling Tool Functions with vMF Distributions}
\label{sec:method:vmf}

Our core idea is to model each function class as a compact directional distribution, allowing calls from different domains to share a common direction whenever they implement the same function.

\textbf{vMF model.}
For each function class $k\in\mathcal{C}$, we model its normalized representations with a vMF distribution, which is analogous to spherical Gaussian distributions for embeddings with unit norms:
\begin{equation}
p(\mathbf{z}\mid c=k)
=
C_d(\beta)
\exp\big(\beta\boldsymbol{\mu}_k^\top\mathbf{z}\big),
\label{eq:vmf}
\end{equation}
where $\boldsymbol{\mu}_k\in\mathbb{S}^{d-1}$ is the prototype direction, $\beta\geq0$ is the concentration, and $C_d(\beta)$ is the normalizing constant. A larger $\beta$ concentrates the distribution more tightly around $\boldsymbol{\mu}_k$, whereas $\beta=0$ yields a uniform distribution on the hypersphere.

Under this probability model, the posterior probability of function class $k$ is
\begin{equation}
p(c=k\mid\mathbf{z})
=\frac{C_d(\beta)\exp(\beta\boldsymbol{\mu}_k^\top\mathbf{z})}
{\sum_{j\in\mathcal{C}}C_d(\beta)\exp(\beta\boldsymbol{\mu}_j^\top\mathbf{z})}
=\frac{\exp(\boldsymbol{\mu}_k^\top\mathbf{z}/\tau)}
{\sum_{j\in\mathcal{C}}\exp(\boldsymbol{\mu}_j^\top\mathbf{z}/\tau)},
\qquad \beta=\frac{1}{\tau}.
\label{eq:posterior}
\end{equation}
where $\tau>0$ is the temperature. Since $\boldsymbol{\mu}_k$ and $\mathbf{z}$ have unit norm, their inner product is the cosine similarity. A smaller $\tau$, equivalently a larger $\beta$, produces a more concentrated function distribution.

\textbf{Reducing intra-function variation.} Consider a minibatch of tool calls
$\mathcal{B}=\{(\mathbf{z}_i,c_i)\}_{i=1}^{N}$, where $c_i=c(u_i)$ is the function class of the called training tool. We minimize the negative log-likelihood of the observed function assignments:
\begin{equation}
\mathcal{L}_{\mathrm{var}}
=
-\frac{1}{N}
\sum_{i=1}^{N}
\log p(c=c_i\mid\mathbf{z}_i).
\label{eq:var}
\end{equation}
Minimizing $\mathcal{L}_{\mathrm{var}}$ aligns each tool call with its function prototype, thereby reducing intra-function variation across tools, domains, and interaction contexts.

\textbf{Increasing inter-function separation.}
The softmax denominator in $\mathcal{L}_{\mathrm{var}}$ already contrasts calls with competing prototypes; $\mathcal{L}_{\mathrm{sep}}$ adds class-level separation. For each function class represented in the minibatch, we compute its normalized batch direction
$
\bar{\mathbf{z}}_k
=
\sum_{i\in\mathcal{I}_k}\mathbf{z}_i / 
\left\lVert\sum_{i\in\mathcal{I}_k}\mathbf{z}_i\right\rVert_2,
\mathcal{I}_k
=
\{\,i:c_i=k\,\}, 
 k\in\mathcal{C}_{\mathcal{B}},
$
where $\mathcal{C}_{\mathcal{B}}=\{c_i:(\mathbf{z}_i,c_i)\in\mathcal{B}\}$ denotes the function classes appearing in the minibatch. We then define
\begin{equation}
\mathcal{L}_{\mathrm{sep}}
=
\frac{1}{|\mathcal{C}_{\mathcal{B}}|}
\sum_{k\in\mathcal{C}_{\mathcal{B}}}
\log
\left[
\frac{1}{|\mathcal{C}|-1}
\sum_{\substack{j\in\mathcal{C}, j\neq k}}
\exp\left(
\frac{\bar{\mathbf{z}}_k^\top\boldsymbol{\mu}_j}{\tau}
\right)
\right].
\label{eq:sep}
\end{equation}
Minimizing $\mathcal{L}_{\mathrm{sep}}$ pushes each batch direction away from the prototypes of other function classes, increasing their angular separation. The prototypes are treated as stop-gradient targets when computing both losses, so $\mathcal{L}_{\mathrm{var}}$ and $\mathcal{L}_{\mathrm{sep}}$ update the policy and projection head through the current representations.

\textbf{Prototype estimation.}
We maintain each function prototype using an exponential moving average. After processing a minibatch, the prototype of every observed class is updated as
\begin{equation}
\boldsymbol{\mu}_k
\leftarrow
\operatorname{Normalize}\left(
\alpha\boldsymbol{\mu}_k
+(1-\alpha)
\bar{\mathbf{z}}_k
\right),
\qquad k\in\mathcal{C}_{\mathcal{B}},
\label{eq:ema}
\end{equation}
where $\alpha\in[0,1)$ is the update factor. The prototypes  track the evolving tool-call representations.

\textbf{Representation-shaping objective.}
The complete objective is
\begin{equation}
\mathcal{L}_{\mathrm{vMF}}
=
\mathcal{L}_{\mathrm{var}}
+
\lambda_{\mathrm{sep}}\mathcal{L}_{\mathrm{sep}},
\label{eq:vmf-objective}
\end{equation}
where $\lambda_{\mathrm{sep}}$ is the weight coefficient modulating the relative importance of the two losses. The two losses organize tool-call representations into compact and separated function clusters. This structure makes knowledge learned from a seen tool transferable to unseen tools implementing the same function, directing trialing toward functionally plausible tools rather than unrelated alternatives.

\Needspace{18\baselineskip}
\begin{wrapfigure}{r}{0.43\textwidth}
\centering

\includegraphics[width=\linewidth,trim=0 0 0 0,clip]{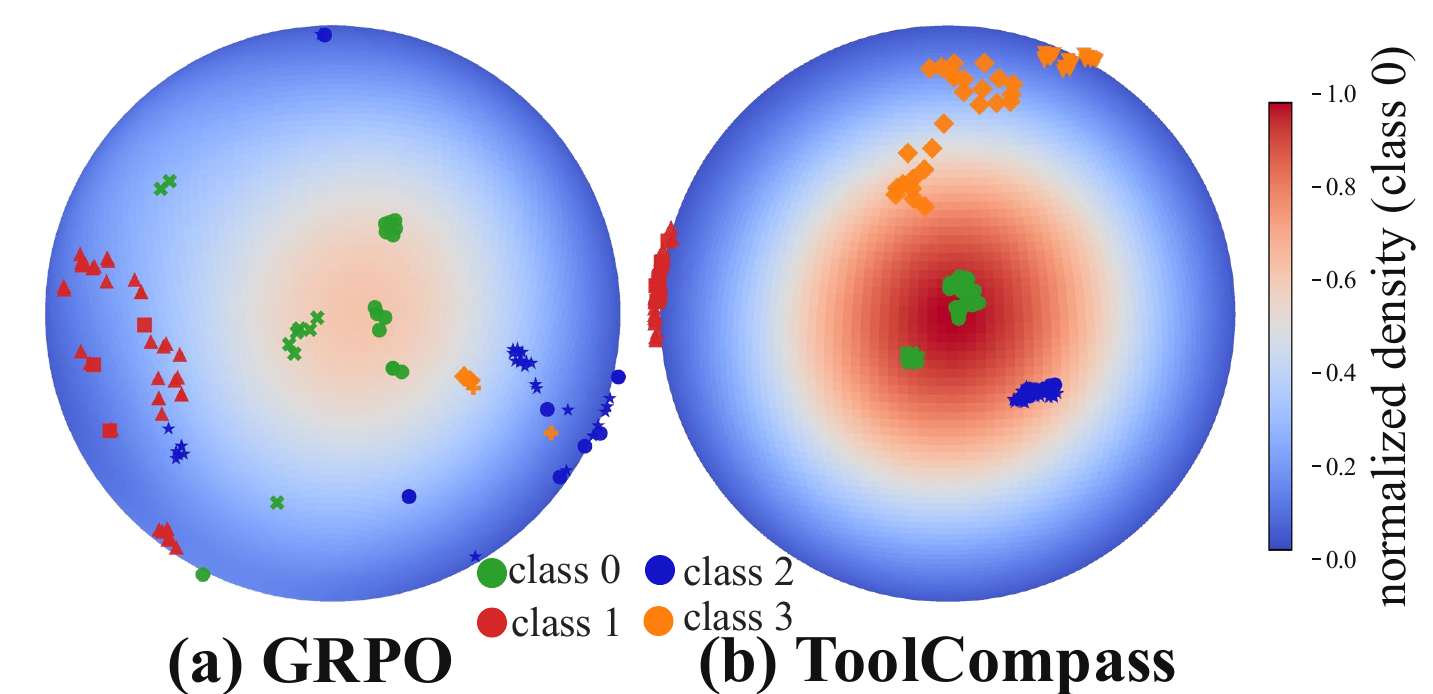}

\caption{ The surface shows the spherical kernel density of one function class (class 0) under GRPO (a) and \method (b), on a shared color scale. We showcase four classes for visual clarity; colors denote function classes and markers denote tools.}
\label{fig:vmf_sphere}
\vspace{-0.3cm}
\end{wrapfigure}

\textbf{Comparison with existing methods.}
Outcome-based methods such as GRPO apply the same trajectory-level advantage to the tool calls within a rollout, indicating whether the trajectory succeeds but not what function each call performs. In contrast, $\mathcal{L}_{\mathrm{vMF}}$ provides function-level supervision: calls implementing the same function share a directional target even when they involve different tools or domains. Unlike trace-matching supervision~\citep{qu2026matchtir}, \method does not prescribe a gold sequence of calls; unlike turn-contribution methods~\citep{tao2026trace,xie2026tips}, it does not require estimating the contribution of individual turns. It therefore complements outcome supervision by guiding where the agent trials without directly penalizing exploration. Figure~\ref{fig:vmf_sphere} visualizes the normalized representation space, with the spherical surface showing the kernel density of a representative function class; compared with GRPO's diffuse distribution, \method forms a compact cluster across tools (see Appendix~\ref{app:vis_details}).

\subsection{Joint Optimization with Post-Training}
\label{sec:method:joint}

We optimize the representation-shaping objective jointly with the original post-training objective:
\begin{equation}
\mathcal{L}(\theta,\psi)
=
\mathcal{L}_{\mathrm{post}}(\theta)
+
{\lambda_{\mathrm{vMF}}}\mathcal{L}_{\mathrm{vMF}}(\theta,\psi).
\label{eq:joint}
\end{equation}
Here, $\mathcal{L}_{\mathrm{post}}$ uses task feedback to learn which tool-use behaviors help complete the task. $\mathcal{L}_{\mathrm{vMF}}$ aligns call representations across tools that perform the same function. Joint training encourages the policy to transfer useful tool-use experience from seen tools to unseen tools with matching functions. The host objective can be instantiated with GRPO~\citep{shao2024deepseekmath}, RFT~\citep{yuan2023rft}, or DMPO~\citep{shi2024dmpo} without changing its original formulation. The overall procedure is provided in Algorithm~\ref{alg} in Appendix~\ref{app}. 

Our  training components consist only of the projection head $g_\psi$ and one prototype per function class. Both are not utilized at deployment. Since $\mathcal{L}_{\mathrm{vMF}}$ has already shaped the policy parameters $\theta$, the agent uses its original policy without an auxiliary scoring module or additional inference pass.

\section{Experiments}
\label{sec:experiments}

\subsection{Setup}
\label{sec:exp-setup}

\textbf{Benchmarks, models, and evaluation.}
We evaluate \method on AppWorld~\citep{trivedi2024appworld} and FTRL~\citep{ye2026feedback}, two multi-turn tool-use benchmarks with executable tools and verifiable task outcomes. We additionally evaluate semantic OOD transfer to three jointly held-out function classes on FTRL (Appendix~\ref{app:semantic-ood}). In the main experiments, we partition the available tools into a seen set $\mathcal{U}_{\mathrm{train}}$ and a held-out set $\mathcal{U}_{\mathrm{unseen}}$. ID tasks can be completed using tools from $\mathcal{U}_{\mathrm{train}}$, whereas OOD tasks require at least one tool from $\mathcal{U}_{\mathrm{unseen}}$. The unseen tools and their function annotations are inaccessible during post-training; their specifications are provided to the agent only at evaluation time. The AppWorld test set contains 168 ID and 417 OOD tasks, and the FTRL test set contains 168 ID and 32 OOD tasks. We use Qwen3.5-4B and Qwen3.5-9B~\citep{qwen35blog} as policy backbones. Throughout all experiments, we adopt the ReAct interaction scaffold~\citep{yao2023react}, in which the agent interleaves language reasoning with executable tool calls and revises its subsequent decisions based on environment observations, including execution results and error messages. Following AppWorld~\citep{trivedi2024appworld}, we use its state-based evaluator and report Task Success Rate, defined as the percentage of tasks for which all task-specific evaluation tests are passed. Following MatchTIR~\citep{qu2026matchtir}, we report Solve-F1 on FTRL, which balances tool-invocation precision and task-completion recall. For both benchmarks, we report performance on the ID subset, the OOD subset, and the entire test set. Detailed benchmark versions, post-training data, ID/OOD split construction, function-class annotations, prompts, interaction protocols, metric implementations, training and inference configurations, and computational resources are provided in Appendix~\ref{app:experimental-details}.

\textbf{Baselines.}
We compare \method against six categories of baselines. First, prompting-based agents include GPT-5.5~\citep{openai2026gpt55}, Claude Opus 4.8~\citep{anthropic2026claudeopus48}, GLM-5.2~\citep{zeng2026glm}, and DeepSeek V4 Pro~\citep{xu2026deepseek}. These agents use the same ReAct scaffold~\citep{yao2023react}, tool specifications, interaction budget, and evaluation protocol described above. Second, general post-training objectives include GRPO~\citep{shao2024deepseekmath}, RFT~\citep{yuan2023rft}, and DMPO~\citep{shi2024dmpo}; we additionally augment each objective with \method to evaluate its compatibility with different host objectives. Unless otherwise stated, \method uses GRPO as the base objective in all remaining experiments. Third, turn-level tool-use post-training methods include StepTool~\citep{yu2025steptool}, FTRL-M~\citep{ye2026feedback}, MatchTIR~\citep{qu2026matchtir}, SOAR~\citep{li2026soar}, and TRACE~\citep{tao2026trace}. Fourth, tool-use RL methods include SimpleTIR~\citep{xue2026simpletir}, ToolMaster~\citep{gao2026teaching}, and LOOP~\citep{chen2025reinforcement}. Fifth, we include SEAL~\citep{li2026cyclical} as a representation-learning baseline. Finally, the OOD-generalization comparison includes CORAL~\citep{sun2016deep}, Group DRO~\citep{sagawa2020distributionally}, ToolRL~\citep{qian2026toolrl}, and PAFT~\citep{lv2026can}. All trainable baselines use the same policy initialization, post-training data, tool specifications, and interaction budget whenever applicable. Baseline-specific adaptations and reproduction details are provided in Appendix~\ref{app:baselines}.

\subsection{Main Results}
\label{sec:main-results}

\begin{table}[t]

  \centering
  \caption{ \textbf{\method\ across post-training objectives.}
  Results are task success rate (\%) on AppWorld and Solve-F1 (\%) on FTRL. Each objective is compared with its
  \method-augmented counterpart under the same backbone and training setup.}
  \label{tab:optimizer-generality}
\fontsize{9}{10.5}\selectfont
  \setlength{\tabcolsep}{6.5pt}
  \renewcommand{\arraystretch}{1.05}
  \setlength{\aboverulesep}{1pt}
  \setlength{\belowrulesep}{1.5pt}
  \begin{tabular*}{\textwidth}{@{\extracolsep{\fill}}llrrrrrr@{}}
    \toprule
    & & \multicolumn{3}{c}{AppWorld}
        & \multicolumn{3}{c}{FTRL} \\
    \cmidrule(lr){3-5}\cmidrule(lr){6-8}
    Method & Variant & ID & OOD & Total & ID & OOD & Total \\
    \midrule
    \rowcolor{hypogainsboro}
    \multicolumn{8}{c}{\textit{\textbf{Qwen3.5-4B}}} \\
    \multirow{2}{*}{GRPO~\citep{shao2024deepseekmath}}
      & Original
      & 58.33 & 56.83 & 57.26
      & 34.82 & 60.40 & 38.91 \\
      & \ \ \textbf{+\method}
      & \textbf{70.63} & \textbf{64.75} & \textbf{66.44}
      & \textbf{46.97} & \textbf{64.01} & \textbf{49.70} \\
    \cmidrule(lr){1-8}
    \multirow{2}{*}{RFT~\citep{yuan2023rft}}
      & Original
      & 44.64 & 35.01 & 37.78
      & 34.93 & 52.17 & 37.69 \\
      & \ \ \textbf{+\method}
      & \textbf{46.43} & \textbf{39.09} & \textbf{41.20}
      & \textbf{40.98} & \textbf{55.78} & \textbf{43.35} \\
    \cmidrule(lr){1-8}
    \multirow{2}{*}{DMPO~\citep{shi2024dmpo}}
      & Original
      & 39.29 & 30.94 & 33.33
      & 36.79 & 45.68 & 38.21 \\
      & \ \ \textbf{+\method}
      & \textbf{44.44} & \textbf{33.97} & \textbf{36.98}
      & \textbf{42.37} & \textbf{52.92} & \textbf{44.06} \\
    \midrule
    \rowcolor{hypogainsboro}
    \multicolumn{8}{c}{\textit{\textbf{Qwen3.5-9B}}} \\
    \multirow{2}{*}{GRPO~\citep{shao2024deepseekmath}}
      & Original
      & 70.83 & 61.15 & 63.93
      & 38.70 & 61.20 & 42.30 \\
      & \ \ \textbf{+\method}
      & \textbf{78.17} & \textbf{70.02} & \textbf{72.36}
      & \textbf{48.99} & \textbf{64.98} & \textbf{51.55} \\
    \cmidrule(lr){1-8}
    \multirow{2}{*}{RFT~\citep{yuan2023rft}}
      & Original
      & 55.36 & 40.05 & 44.44
      & 35.94 & 46.70 & 37.66 \\
      & \ \ \textbf{+\method}
      & \textbf{60.12} & \textbf{50.76} & \textbf{53.45}
      & \textbf{40.70} & \textbf{57.39} & \textbf{43.37} \\
    \cmidrule(lr){1-8}
    \multirow{2}{*}{DMPO~\citep{shi2024dmpo}}
      & Original
      & 60.71 & 49.16 & 52.48
      & 37.63 & 53.93 & 40.24 \\
      & \ \ \textbf{+\method}
      & \textbf{65.67} & \textbf{55.32} & \textbf{58.29}
      & \textbf{46.90} & \textbf{55.73} & \textbf{48.31} \\
    \bottomrule
  \end{tabular*}
  
\end{table}

\begin{figure}[t]
    
    \centering
    \includegraphics[width=0.86\textwidth]{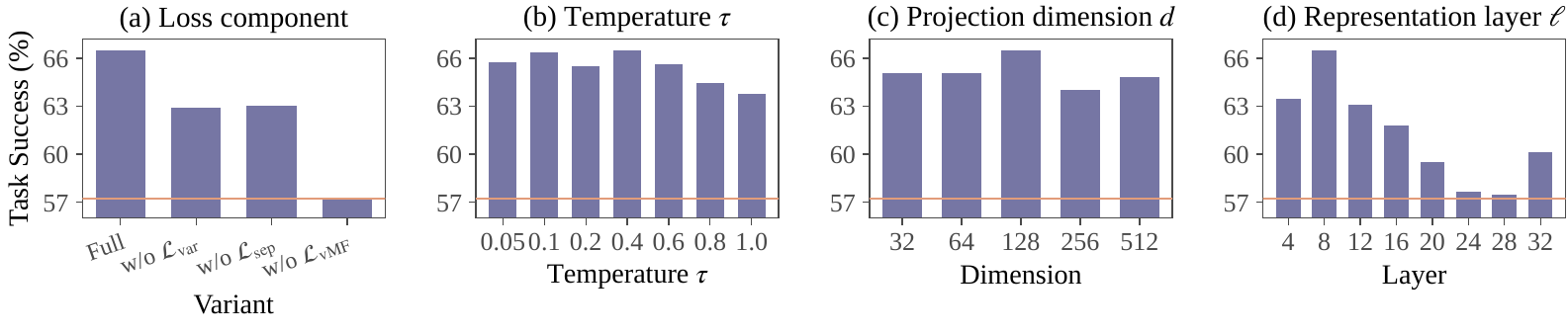}
    
    \caption{ \textbf{Ablation studies.}
    (a) Effect of loss components, (b) effect of the vMF temperature $\tau$,
    (c) effect of projection dimension $d$, and
    (d) effect of representation layer $l$.
    Results are AppWorld task success rates using Qwen3.5-4B with GRPO
    and are averaged over three seeds.
    The orange horizontal line denotes vanilla GRPO.}
    \label{fig:analysis-ablation}
    
\end{figure}

\textbf{\method benefits across post-training objectives.}
Table~\ref{tab:optimizer-generality} compares each post-training objective with its \method-augmented counterpart. \method improves every reported ID, OOD, and total score across both benchmarks and model scales. On Qwen3.5-4B, it improves the AppWorld and FTRL total scores by 9.18 and 10.79 percentage points with GRPO, 3.42 and 5.66 points with RFT, and 3.65 and 5.85 points with DMPO, respectively. The improvements remain consistent on Qwen3.5-9B, reaching up to 9.01 points on AppWorld and 9.25 points on FTRL. These results demonstrate that \method is consistently effective across different post-training objectives and model scales.

\textbf{Comparison with competitive baselines.}
Table~\ref{tab:main-comparison} compares \method with prompting-based agents and trainable tool-use methods. \method achieves the best OOD and total scores across both benchmarks and model scales. On AppWorld with Qwen3.5-9B, it reaches a total task success rate of 72.36\%, compared with 68.26\% for LOOP~\citep{chen2025reinforcement}, the strongest baseline on this metric, a gain of 4.10 percentage points. We note that a few proprietary models remain competitive on the AppWorld ID split, likely because \textit{their training corpora may already cover related applications and tool-use data}, yet \method surpasses all prompting-based agents on every OOD and total score.

\textbf{Comparison with OOD-generalization methods.}
Table~\ref{tab:ood-comparison} compares \method with general and agent-oriented OOD-generalization methods under GRPO with Qwen3.5-4B. On AppWorld, \method achieves an OOD score of 64.75\% and a total score of 66.44\%, exceeding CORAL, the strongest baseline on both metrics, by 6.48 and 6.27 percentage points, respectively. On FTRL, \method reaches an OOD score of 64.01\% and a total score of 49.70\%, exceeding GRPO by 3.61 points on OOD and ToolRL by 7.12 points on total performance. \method thus generalizes to tasks requiring unseen tools more effectively than the compared baselines.

\begin{table}[t]
\caption{ \textbf{Main comparison on AppWorld and FTRL.} Results are task success rate (\%) on AppWorld and Solve-F1 (\%) on FTRL. All prompting models use the same ReAct scaffold~\citep{yao2023react}. Rep.\ learning abbreviates representation learning, and FTRL-M is the multi-turn variant of the FTRL method. For brevity, we report mean $\pm$ standard deviation over the same three seeds for LOOP, SEAL, and \method.}
\label{tab:main-comparison}

\centering
\fontsize{9}{10.5}\selectfont
\setlength{\tabcolsep}{2.5pt}
\renewcommand{\arraystretch}{1.05}
\begin{tabular*}{\textwidth}{@{\extracolsep{\fill}}llrrrrrr@{}}
\toprule
& & \multicolumn{3}{c}{AppWorld} & \multicolumn{3}{c}{FTRL} \\
\cmidrule(lr){3-5}\cmidrule(lr){6-8}
Method & Category & ID & OOD & Total & ID & OOD & Total \\
\midrule
\rowcolor{hypogainsboro}
    \multicolumn{8}{c}{\textit{\textbf{Prompting-based}}}  \\
GPT-5.5~\citep{openai2026gpt55} & Prompting & 70.24 & 63.07 & 65.13 & 33.06 & 54.32 & 36.46 \\
Claude Opus 4.8~\citep{anthropic2026claudeopus48} & Prompting & 70.83 & 58.51 & 62.05 & 44.18 & 49.64 & 45.05 \\
GLM-5.2~\citep{zeng2026glm} & Prompting & 58.33 & 54.92 & 55.90 & 40.64 & 43.85 & 41.15 \\
DeepSeek V4 Pro~\citep{xu2026deepseek} & Prompting & 70.24 & 55.64 & 59.83 & 35.52 & 52.23 & 38.20 \\

    \midrule
    \rowcolor{hypogainsboro}
    \multicolumn{8}{c}{
      \textit{\textbf{Qwen3.5-4B}}
    }  \\
GRPO~\citep{shao2024deepseekmath} & Base RL & 58.33 & 56.83 & 57.26 & 34.82 & 60.40 & 38.91 \\
DMPO~\citep{shi2024dmpo} & Base RL & 39.29 & 30.94 & 33.33 & 36.79 & 45.68 & 38.21 \\
StepTool~\citep{yu2025steptool} & Turn-level RL & 65.48 & 52.76 & 56.41 & 40.11 & 49.59 & 41.63 \\
FTRL-M~\citep{ye2026feedback} & Turn-level RL & 65.48 & 53.00 & 56.58 & 36.81 & 59.01 & 40.36 \\
MatchTIR~\citep{qu2026matchtir} & Turn-level RL & 69.05 & 53.96 & 58.29 & 39.91 & 50.41 & 41.59 \\
SOAR~\citep{li2026soar} & Turn-level RL & 64.88 & 50.12 & 54.36 & 35.55 & 57.64 & 39.08 \\
TRACE~\citep{tao2026trace} & Turn-level RL & 69.05 & 55.16 & 59.15 & 39.26 & 55.94 & 41.93 \\
SimpleTIR~\citep{xue2026simpletir} & Tool RL & 63.69 & 55.64 & 57.95 & 42.94 & 55.45 & 44.95 \\
ToolMaster~\citep{gao2026teaching} & Tool RL & 60.12 & 57.55 & 58.29 & 42.85 & 48.26 & 43.71 \\
LOOP~\citep{chen2025reinforcement} & Tool RL & \ms{65.87}{0.43} & \ms{58.99}{0.92} & \ms{60.97}{0.65} & \ms{41.83}{1.03} & \ms{57.72}{2.14} & \ms{44.37}{1.07} \\
SEAL~\citep{li2026cyclical} & Rep.\ learning & \ms{67.26}{1.19} & \ms{59.23}{0.96} & \ms{61.54}{1.03} & \ms{38.02}{1.49} & \ms{57.94}{2.51} & \ms{41.21}{1.63} \\
\textbf{\method (ours)} & \textbf{Rep.\ learning} & \textbf{\boldmath\ms{70.63}{0.91}} & \textbf{\boldmath\ms{64.75}{1.27}} & \textbf{\boldmath\ms{66.44}{1.09}} & \textbf{\boldmath\ms{46.97}{1.76}} & \textbf{\boldmath\ms{64.01}{2.39}} & \textbf{\boldmath\ms{49.70}{1.53}} \\

    \midrule
    \rowcolor{hypogainsboro}
    \multicolumn{8}{c}{
      \textit{\textbf{Qwen3.5-9B}}
    }  \\
GRPO~\citep{shao2024deepseekmath} & Base RL & 70.83 & 61.15 & 63.93 & 38.70 & 61.20 & 42.30 \\
DMPO~\citep{shi2024dmpo} & Base RL & 60.71 & 49.16 & 52.48 & 37.63 & 53.93 & 40.24 \\
StepTool~\citep{yu2025steptool} & Turn-level RL & 74.40 & 59.95 & 64.10 & 46.50 & 44.23 & 46.13 \\
FTRL-M~\citep{ye2026feedback} & Turn-level RL & 72.02 & 61.63 & 64.62 & 42.79 & 51.36 & 44.16 \\
MatchTIR~\citep{qu2026matchtir} & Turn-level RL & 75.60 & 58.27 & 63.25 & 48.41 & 45.36 & 47.93 \\
SOAR~\citep{li2026soar} & Turn-level RL & 62.50 & 53.24 & 55.90 & 42.85 & 52.28 & 44.36 \\
TRACE~\citep{tao2026trace} & Turn-level RL & 74.40 & 57.79 & 62.56 & 48.89 & 49.92 & 49.05 \\
SimpleTIR~\citep{xue2026simpletir} & Tool RL & 72.02 & 65.71 & 67.52 & 46.04 & 58.27 & 47.99 \\
ToolMaster~\citep{gao2026teaching} & Tool RL & 71.43 & 61.87 & 64.62 & 44.67 & 54.96 & 46.31 \\
LOOP~\citep{chen2025reinforcement} & Tool RL & \ms{71.03}{0.57} & \ms{67.15}{0.91} & \ms{68.26}{0.71} & \ms{43.61}{1.27} & \ms{58.55}{1.41} & \ms{46.00}{0.99} \\
SEAL~\citep{li2026cyclical} & Rep.\ learning & \ms{76.79}{1.19} & \ms{63.47}{0.97} & \ms{67.29}{1.03} & \ms{46.92}{1.48} & \ms{58.53}{2.46} & \ms{48.78}{1.57} \\
\textbf{\method (ours)} & \textbf{Rep.\ learning} & \textbf{\boldmath\ms{78.17}{1.37}} & \textbf{\boldmath\ms{70.02}{1.87}} & \textbf{\boldmath\ms{72.36}{1.55}} & \textbf{\boldmath\ms{48.99}{1.21}} & \textbf{\boldmath\ms{64.98}{2.18}} & \textbf{\boldmath\ms{51.55}{1.07}} \\

    \bottomrule
  
\end{tabular*}
\end{table}

\subsection{Analysis}
\label{sec:exp:analysis}

\textbf{Effect of loss components.}
We ablate each loss component by removing it from the full objective. As shown in Figure~\ref{fig:analysis-ablation} (a), removing any component degrades performance. Removing $\mathcal{L}_{\mathrm{vMF}}$ causes the largest drop, from 66.44\% to 57.21\%, as the policy no longer receives function-level representation shaping. Removing $\mathcal{L}_{\mathrm{var}}$ allows tool calls implementing the same function to spread apart, while removing $\mathcal{L}_{\mathrm{sep}}$ reduces the separation between different function prototypes. Both losses contribute to the performance of the full objective.

\Needspace{16\baselineskip}
\begin{wrapfigure}{r}{0.48\textwidth}
    
    \centering
    \includegraphics[width=0.98\linewidth]{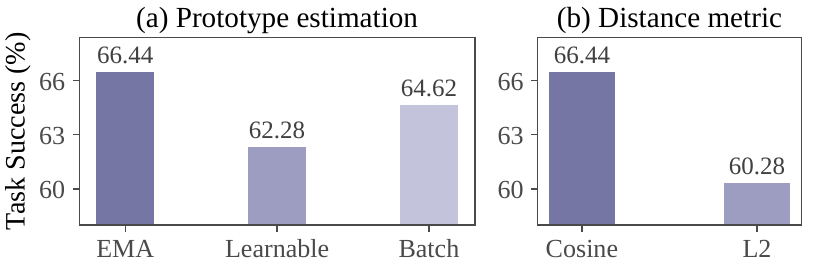}
    
    \caption{ \textbf{Analysis of design choices.}
    (a) Effect of prototype estimation and
    (b) effect of distance metric.
    Results are AppWorld task success rates using Qwen3.5-4B with GRPO
    and are averaged over three seeds.}
    \label{fig:design-ablation}
    
\end{wrapfigure}

\textbf{Effect of the temperature $\tau$.}
We vary the temperature $\tau$ of the vMF objective. Figure~\ref{fig:analysis-ablation} (b) shows that performance peaks at $\tau=0.4$ and remains competitive across a wide range, suggesting that \method is robust to the temperature setting.

\textbf{Effect of the output dimension $d$.}
We vary the output dimension of the projection head. As shown in Figure~\ref{fig:analysis-ablation} (c), performance peaks at $d=128$ and remains competitive across a wide range, showing that \method is not sensitive to a specific projection dimension.

\textbf{How do different layers impact \method?}
In Figure~\ref{fig:analysis-ablation} (d), we extract tool-call representations from different layers. Performance peaks at layer $l=8$, suggesting that early-to-intermediate layers preserve the most transferable function-level information. \method remains above vanilla GRPO across all tested layers, indicating that it does not rely on a specific layer depth to be effective. Performance at the top layers degrades, likely because their representations become specialized for next-token prediction and drift away from the function-level semantics of the call.

\suppressfloats[t]
\begin{table}[t]
    
    \centering
  \caption{ \textbf{Comparison with OOD generalization methods.}
  All methods use GRPO with Qwen3.5-4B. Results are task success rate (\%) on AppWorld and Solve-F1 (\%) on FTRL.}
  \label{tab:ood-comparison}
\normalsize
  \setlength{\tabcolsep}{4pt}
  \renewcommand{\arraystretch}{1.02}
  \begin{tabular*}{\textwidth}{@{\extracolsep{\fill}}llrrrrrr@{}}
    \toprule
    & & \multicolumn{3}{c}{AppWorld}
    & \multicolumn{3}{c}{FTRL} \\
    \cmidrule(lr){3-5}\cmidrule(lr){6-8}
    Method & Category & ID & OOD & Total & ID & OOD & Total \\
    \midrule
    GRPO~\citep{shao2024deepseekmath}
      & Vanilla RL
      & 58.33 & 56.83 & 57.26
      & 34.82 & 60.40 & 38.91 \\
    CORAL~\citep{sun2016deep}
      & General OOD
      & 64.88 & 58.27 & 60.17
      & 33.73 & 58.98 & 37.77 \\
    Group DRO~\citep{sagawa2020distributionally}
      & General OOD
      & 67.86 & 54.68 & 58.46
      & 35.55 & 58.98 & 39.30 \\
    ToolRL~\citep{qian2026toolrl}
      & Agentic OOD
      & 68.45 & 56.12 & 59.66
      & 39.20 & 60.32 & 42.58 \\
    PAFT~\citep{lv2026can}
      & Agentic OOD
      & 60.12 & 51.56 & 54.02
      & 36.47 & 57.84 & 39.89 \\
    \midrule
    \textbf{\method (ours)}
      & \textbf{Agentic OOD}
      & \textbf{70.63} & \textbf{64.75} & \textbf{66.44}
      & \textbf{46.97} & \textbf{64.01} & \textbf{49.70} \\
    \bottomrule
  \end{tabular*}
  
\end{table}

\textbf{Effect of prototype estimation.}
We compare EMA prototypes with learnable and batch-estimated prototypes.
As shown in Figure~\ref{fig:design-ablation}(a), EMA prototype estimation
achieves the best performance, suggesting that smoothly tracking the
evolving representation space provides more stable function-level targets. We further ablate the EMA factor $\alpha$ in Appendix~\ref{app:ema-ablation}.

\textbf{Effect of distance metric.}
We compare cosine similarity with negative unsquared L2 scores at a fixed temperature (Appendix~\ref{app:implementation-details}). Figure~\ref{fig:design-ablation}(b) shows that cosine achieves 66.44\% task success, compared with 60.28\% for L2 under the same configuration.

\begin{figure}[t]
    
    \centering
    \includegraphics[width=0.90\linewidth]{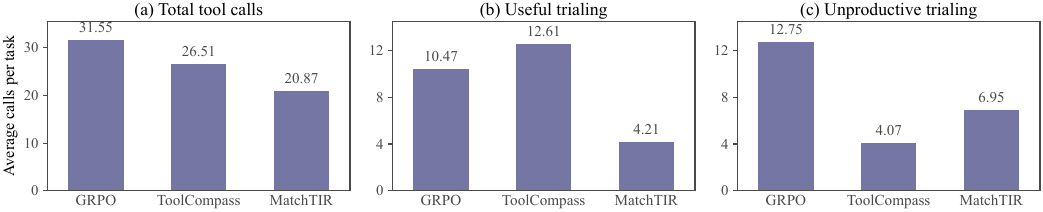}
    
    \caption{ \textbf{Tool-call analysis on AppWorld OOD tasks.}
    (a) Total tool calls, (b) useful trialing, and
    (c) unproductive trialing.
    \method shows more useful trialing and less unproductive trialing than GRPO.}
    \label{fig:tool-trialing}
    
\end{figure}

\Needspace{15\baselineskip}
\begin{wraptable}{r}{0.45\textwidth}
\vspace{-\intextsep}
    
    \centering
\small
    \caption{ \textbf{Generalization across datasets.} Results (\%) compare vanilla GRPO and \method using Qwen3.5-4B.}
    \label{tab:cross-dataset}
\small
    \setlength{\tabcolsep}{4pt}
    \renewcommand{\arraystretch}{1.05}
    \begin{tabular*}{\linewidth}{@{\extracolsep{\fill}}lcc@{}}
        \toprule
        Test dataset & GRPO & \method \\
        \midrule
        \multicolumn{3}{@{}l}{\textit{Training dataset: AppWorld}} \\
        \addlinespace[2pt]
        AppWorld & 57.21 & \textbf{66.44} \\
        FTRL     & 35.72 & \textbf{42.31} \\
        \midrule
        \multicolumn{3}{@{}l}{\textit{Training dataset: FTRL}} \\
        \addlinespace[2pt]
        AppWorld & 53.03 & \textbf{60.51} \\
        FTRL     & 38.91 & \textbf{49.70} \\
        \bottomrule
    \end{tabular*}

\end{wraptable}

\textbf{Generalization across datasets.}
We train GRPO and \method on a source dataset and directly evaluate them on a different target dataset. As shown in Table~\ref{tab:cross-dataset}, \method outperforms vanilla GRPO in both transfer directions. For example, training \method on FTRL yields 60.51\% task success on AppWorld, compared with 66.44\% when trained on AppWorld itself. These results suggest that the benefits of \method extend to cross-dataset transfer.

\textbf{Does \method guide rather than suppress tool trialing?}
We categorize tool calls following Appendix~\ref{app:tool-call-categorization}. Figure~\ref{fig:tool-trialing} shows that \method makes fewer calls than GRPO (26.51 vs.\ 31.55) but more than MatchTIR (20.87). It retains more useful trials than GRPO and MatchTIR (12.61 vs.\ 10.47 and 4.21) and fewer unproductive trials than GRPO (4.07 vs.\ 12.75). These results are consistent with guided tool trialing: \method retains more useful trials while reducing unproductive ones.

\begingroup\raggedright
\textbf{Additional analysis.} We provide further analyses in the Appendix: (1) semantic OOD analysis on FTRL with three jointly held-out function classes (Appendix~\ref{app:semantic-ood}); (2) comparisons with alternative training objectives (Appendix~\ref{app:auxiliary-objectives}); (3) ablations on loss weights (Appendix~\ref{app:loss-weights}) and the EMA factor $\alpha$ in Equation~\ref{eq:ema} (Appendix~\ref{app:ema-ablation}); (4) qualitative case studies on AppWorld OOD tasks (Appendix~\ref{app:qualitative}); and (5) AppWorld metrics on Scenario Goal Completion (Appendix~\ref{app:sgc}).

\par\endgroup

\section{Related Work}
\label{sec:related}

\textbf{OOD generalization.}
Conventional OOD generalization improves robustness through domain alignment, robust optimization, or  representation learning~\citep{sun2016deep,sagawa2020distributionally}. These methods study fixed prediction, whereas tool-use agents face shifts in tasks, tools, observations, and interactions. Recent studies expose weak transfer across agent environments~\citep{xi2026can,lv2026can}, while tool-specific work uses synthetic environments~\citep{tang2023toolalpaca,he2025gentool,sullivan2025procedural,du2026generalizable}, self-verification or interface optimization~\citep{mekala2024toolverifier,guo2026learning}, open-world retrieval~\citep{wu2025chain,molfetta2025ports,zhuang2026losemb,huang2026toolomni}, generalization-aware rewards~\citep{zeng2025tool,qian2026toolrl}, and documentation adaptation for evolving tools~\citep{wu2026beyond}. \method instead organizes tool-call representations by shared function during post-training, enabling transfer to unseen tools without accessing them or adding test-time adaptation.

\textbf{Tool-use agents.}
LLM agents use external tools through prompting, pretraining, and instruction tuning~\citep{yao2023react,schick2023toolformer,shen2023hugginggpt,liang2024taskmatrix,li2023api,patil2024gorilla,qin2024toolllm}. Post-training methods optimize trajectory outcomes~\citep{jin2025searchr1,qian2026toolrl,feng2026retool,chen2025reinforcement,xue2026simpletir,zeng2025tool}, while environment-feedback and teacher-guided methods learn self-correction and trial-and-error~\citep{gou2024critic,shinn2023reflexion,song2024trial,su2026failure}. Sparse long-horizon rewards motivate turn-level and error-localized supervision~\citep{yu2025steptool,zeng2025reinforcing,luo2025agent,li2026turn,qu2026matchtir,tao2026trace,xie2026tips,liang2026learning}, which matches gold traces, estimates turn contributions, or localizes irrecoverable actions. ToolMaster~\citep{gao2026teaching} is a related approach to tool trialing; however, it requires an additional teacher model to generate trialing trajectories. \method instead guides trialing through shared-function representations, without additional trajectory supervision, frozen turn scorers, or inference overhead.

\section{Conclusion}
\label{sec:conclusion}

We propose \method, a novel framework for guiding tool trialing to improve OOD tool use. Our framework leverages function-class assignments and a vMF objective to shape the tool-call representation space, enabling experience to transfer from seen tools to functionally related unseen tools. \method requires neither ground-truth call traces nor unseen-tool access, introduces no inference overhead, and can be readily plugged into diverse post-training methods. Extensive experiments show that \method consistently improves tool-use performance across benchmarks, model scales, and post-training methods. We hope our work can inspire future research on guided tool trialing and more broadly on building generalizable tool-use agents in the wild.

\FloatBarrier
\begingroup
\setlength{\bibsep}{1.5pt}
\apptocmd{\thebibliography}{\setlength{\itemsep}{1.5pt}\setlength{\parskip}{0pt}}{}{}
\bibliographystyle{plainnat}
\bibliography{references}

\begin{thebibliography}{59}
\providecommand{\natexlab}[1]{#1}
\providecommand{\url}[1]{\texttt{#1}}
\expandafter\ifx\csname urlstyle\endcsname\relax
  \providecommand{\doi}[1]{doi: #1}\else
  \providecommand{\doi}{doi: \begingroup \urlstyle{rm}\Url}\fi

\bibitem[{Anthropic}(2026)]{anthropic2026claudeopus48}
{Anthropic}.
\newblock {Claude Opus 4.8} system card.
\newblock Technical report, Anthropic, May 2026.

\bibitem[Chen et~al.(2025)Chen, Cusumano-Towner, Huval, Petrenko, Hamburger,
  Koltun, and Kr{\"a}henb{\"u}hl]{chen2025reinforcement}
Kevin Chen, Marco Cusumano-Towner, Brody Huval, Aleksei Petrenko, Jackson
  Hamburger, Vladlen Koltun, and Philipp Kr{\"a}henb{\"u}hl.
\newblock Reinforcement learning for long-horizon interactive llm agents.
\newblock \emph{arXiv preprint arXiv:2502.01600}, 2025.

\bibitem[Du et~al.(2026)Du, Gong, Ling, Liu, Shen, Yao, Xu, Shi, Yang, and
  Chen]{du2026generalizable}
Weihua Du, Hailei Gong, Zhan Ling, Kang Liu, Lingfeng Shen, Xuesong Yao, Yufei
  Xu, Dingyuan Shi, Yiming Yang, and Jiecao Chen.
\newblock Generalizable end-to-end tool-use {RL} with synthetic codegym.
\newblock In \emph{The Fourteenth International Conference on Learning
  Representations}, 2026.
\newblock URL \url{https://openreview.net/forum?id=QRSeFZfu8E}.

\bibitem[Feng et~al.(2026)Feng, Huang, Qu, Zhang, Qin, Zhong, Jiang, Chi, and
  Zhong]{feng2026retool}
Jiazhan Feng, Shijue Huang, Xingwei Qu, Ge~Zhang, Yujia Qin, Baoquan Zhong,
  Chengquan Jiang, Jinxin Chi, and Wanjun Zhong.
\newblock {ReTool}: Reinforcement learning for strategic tool use in {LLMs}.
\newblock In \emph{International Conference on Learning Representations},
  volume 2026, pages 37909--37926, 2026.

\bibitem[Gao et~al.(2026)Gao, Huang, Liu, Yan, Wang, Chen, Qian, Yu, and
  Gu]{gao2026teaching}
Xingjie Gao, Pengcheng Huang, Zhenghao Liu, Yukun Yan, Shuo Wang, Zulong Chen,
  Chen Qian, Ge~Yu, and Yu~Gu.
\newblock Teaching {LLMs} to learn tool trialing and execution through
  environment interaction.
\newblock \emph{arXiv preprint arXiv:2601.12762}, 2026.

\bibitem[Gou et~al.(2024)Gou, Shao, Gong, shen, Yang, Duan, and
  Chen]{gou2024critic}
Zhibin Gou, Zhihong Shao, Yeyun Gong, yelong shen, Yujiu Yang, Nan Duan, and
  Weizhu Chen.
\newblock {CRITIC}: Large language models can self-correct with
  tool-interactive critiquing.
\newblock In \emph{International Conference on Learning Representations},
  volume 2024, pages 57734--57811, 2024.

\bibitem[Guo et~al.(2026)Guo, Dong, Gao, and Das]{guo2026learning}
Ruocheng Guo, Kaiwen Dong, Xiang Gao, and Kamalika Das.
\newblock Learning to rewrite tool descriptions for reliable llm-agent tool
  use.
\newblock \emph{arXiv preprint arXiv:2602.20426}, 2026.

\bibitem[He et~al.(2025)He, Neville, Wan, Yang, Liu, Xu, Song, Pan, and
  Zhou]{he2025gentool}
Jie He, Jennifer Neville, Mengting Wan, Longqi Yang, Hui Liu, Xiaofeng Xu, Xia
  Song, Jeff~Z Pan, and Pei Zhou.
\newblock {GenTool}: Enhancing tool generalization in language models through
  zero-to-one and weak-to-strong simulation.
\newblock In \emph{Findings of the Association for Computational Linguistics:
  ACL 2025}, pages 1097--1122, 2025.

\bibitem[Huang et~al.(2026)Huang, Zhang, Hu, and Zhang]{huang2026toolomni}
Shouzheng Huang, Meishan Zhang, Baotian Hu, and Min Zhang.
\newblock Toolomni: Enabling open-world tool use via agentic learning with
  proactive retrieval and grounded execution.
\newblock In \emph{Proceedings of the 64th Annual Meeting of the Association
  for Computational Linguistics (Volume 1: Long Papers)}, pages 37421--37439,
  2026.

\bibitem[Jin et~al.(2025)Jin, Zeng, Yue, Yoon, Arik, Wang, Zamani, and
  Han]{jin2025searchr1}
Bowen Jin, Hansi Zeng, Zhenrui Yue, Jinsung Yoon, Sercan Arik, Dong Wang, Hamed
  Zamani, and Jiawei Han.
\newblock {Search-R1}: Training {LLMs} to reason and leverage search engines
  with reinforcement learning.
\newblock \emph{arXiv preprint arXiv:2503.09516}, 2025.

\bibitem[Khosla et~al.(2020)Khosla, Teterwak, Wang, Sarna, Tian, Isola,
  Maschinot, Liu, and Krishnan]{khosla2020supervised}
Prannay Khosla, Piotr Teterwak, Chen Wang, Aaron Sarna, Yonglong Tian, Phillip
  Isola, Aaron Maschinot, Ce~Liu, and Dilip Krishnan.
\newblock Supervised contrastive learning.
\newblock \emph{Advances in neural information processing systems},
  33:\penalty0 18661--18673, 2020.

\bibitem[Li et~al.(2026{\natexlab{a}})Li, Zhou, Meng, Vadera, Li, and
  Li]{li2026turn}
Junbo Li, Peng Zhou, Rui Meng, Meet~P Vadera, Lihong Li, and Yang Li.
\newblock Turn-ppo: Turn-level advantage estimation with ppo for improved
  multi-turn rl in agentic llms.
\newblock In \emph{Findings of the Association for Computational Linguistics:
  EACL 2026}, pages 6227--6243, 2026{\natexlab{a}}.

\bibitem[Li et~al.(2026{\natexlab{b}})Li, Li, Wang, Yuan, Wei, Li,
  et~al.]{li2026soar}
Meng Li, Lei Li, Xiting Wang, Yi~Yuan, Zheng Wei, Zang Li, et~al.
\newblock Soar: Supervision from observation for agentic reinforcement
  learning.
\newblock In \emph{Proceedings of the 64th Annual Meeting of the Association
  for Computational Linguistics (Volume 1: Long Papers)}, pages 35175--35197,
  2026{\natexlab{b}}.

\bibitem[Li et~al.(2023)Li, Zhao, Yu, Song, Li, Yu, Li, Huang, and
  Li]{li2023api}
Minghao Li, Yingxiu Zhao, Bowen Yu, Feifan Song, Hangyu Li, Haiyang Yu, Zhoujun
  Li, Fei Huang, and Yongbin Li.
\newblock Api-bank: A comprehensive benchmark for tool-augmented llms.
\newblock In \emph{Proceedings of the 2023 conference on empirical methods in
  natural language processing}, pages 3102--3116, 2023.

\bibitem[Li et~al.(2026{\natexlab{c}})Li, Im, and Li]{li2026cyclical}
Wendi Li, Shawn Im, and Sharon Li.
\newblock Cyclical entropy eruption: Entropy dynamics in agent reinforcement
  learning.
\newblock \emph{arXiv preprint arXiv:2605.27954}, 2026{\natexlab{c}}.

\bibitem[Liang et~al.(2026)Liang, Zhu, Ge, Yang, Shen, Zheng, and
  Guo]{liang2026learning}
Qiao Liang, Yuke Zhu, Chao Ge, Lei Yang, Ying Shen, Bo~Zheng, and Sheng Guo.
\newblock Learning from the irrecoverable: Error-localized policy optimization
  for tool-integrated llm reasoning.
\newblock In \emph{Proceedings of the 64th Annual Meeting of the Association
  for Computational Linguistics (Volume 1: Long Papers)}, pages 11008--11028,
  2026.

\bibitem[Liang et~al.(2024)Liang, Wu, Song, Wu, Xia, Liu, Ou, Lu, Ji, Mao,
  et~al.]{liang2024taskmatrix}
Yaobo Liang, Chenfei Wu, Ting Song, Wenshan Wu, Yan Xia, Yu~Liu, Yang Ou, Shuai
  Lu, Lei Ji, Shaoguang Mao, et~al.
\newblock Taskmatrix. ai: Completing tasks by connecting foundation models with
  millions of apis.
\newblock \emph{Intelligent Computing}, 3:\penalty0 0063, 2024.

\bibitem[Liu et~al.(2025)Liu, Wang, Miao, Hsu, Yan, Chen, Han, Xu, Chen, Jiang,
  et~al.]{liu2025budget}
Tengxiao Liu, Zifeng Wang, Jin Miao, I~Hsu, Jun Yan, Jiefeng Chen, Rujun Han,
  Fangyuan Xu, Yanfei Chen, Ke~Jiang, et~al.
\newblock Budget-aware tool-use enables effective agent scaling.
\newblock \emph{arXiv preprint arXiv:2511.17006}, 2025.

\bibitem[Luo et~al.(2025)Luo, Zhang, He, Wang, Zhao, Li, Qiu, and
  Yang]{luo2025agent}
Xufang Luo, Yuge Zhang, Zhiyuan He, Zilong Wang, Siyun Zhao, Dongsheng Li,
  Luna~K Qiu, and Yuqing Yang.
\newblock Agent lightning: Train any ai agents with reinforcement learning.
\newblock \emph{arXiv preprint arXiv:2508.03680}, 2025.

\bibitem[Lv et~al.(2026)Lv, Wu, Zhu, Cheng, and Guo]{lv2026can}
Song-Lin Lv, Weiming Wu, Rui Zhu, Zi-Jian Cheng, and Lan-Zhe Guo.
\newblock Can agents generalize to the open world? unveiling the fragility of
  static training in tool use.
\newblock \emph{arXiv preprint arXiv:2607.01084}, 2026.

\bibitem[Mansoor et~al.(2026)Mansoor, Phadke, and Rana]{mansoor2026verified}
Isham~Kalappurackal Mansoor, Abhishek Phadke, and Pratip Rana.
\newblock Verified tool calls improve llm agent reliability under non-atomic
  failures.
\newblock \emph{arXiv preprint arXiv:2608.02645}, 2026.

\bibitem[Mardia and Jupp(2009)]{mardia2009directional}
Kanti~V Mardia and Peter~E Jupp.
\newblock \emph{Directional statistics}.
\newblock John Wiley \& Sons, 2009.

\bibitem[Mekala et~al.(2024)Mekala, Weston, Lanchantin, Raileanu, Lomeli,
  Shang, and Dwivedi-Yu]{mekala2024toolverifier}
Dheeraj Mekala, Jason~E Weston, Jack Lanchantin, Roberta Raileanu, Maria
  Lomeli, Jingbo Shang, and Jane Dwivedi-Yu.
\newblock {ToolVerifier}: Generalization to new tools via self-verification.
\newblock In \emph{Findings of the Association for Computational Linguistics:
  EMNLP 2024}, pages 5026--5041, 2024.

\bibitem[Molfetta et~al.(2025)Molfetta, Frisoni, Monaldini, and
  Moro]{molfetta2025ports}
Lorenzo Molfetta, Giacomo Frisoni, Nicol{\`o} Monaldini, and Gianluca Moro.
\newblock Ports: Preference-optimized retrievers for tool selection with large
  language models.
\newblock In \emph{Proceedings of the 2025 Conference on Empirical Methods in
  Natural Language Processing}, pages 10018--10041, 2025.

\bibitem[{OpenAI}(2026)]{openai2026gpt55}
{OpenAI}.
\newblock {GPT-5.5} system card.
\newblock Technical report, OpenAI, April 2026.

\bibitem[Patil et~al.(2024)Patil, Zhang, Wang, and Gonzalez]{patil2024gorilla}
Shishir~G Patil, Tianjun Zhang, Xin Wang, and Joseph~E Gonzalez.
\newblock {Gorilla}: Large language model connected with massive {APIs}.
\newblock \emph{Advances in Neural Information Processing Systems},
  37:\penalty0 126544--126565, 2024.

\bibitem[Qian et~al.(2025)Qian, Acikgoz, He, WANG, Chen, Hakkani-Tur, Tur, and
  Ji]{qian2026toolrl}
Cheng Qian, Emre~Can Acikgoz, Qi~He, Hongru WANG, Xiusi Chen, Dilek
  Hakkani-Tur, Gokhan Tur, and Heng Ji.
\newblock {ToolRL}: Reward is all tool learning needs.
\newblock In \emph{Advances in Neural Information Processing Systems}, volume
  38, Main Conference, pages 105523--105553, 2025.
\newblock \doi{10.52202/085713-3524}.

\bibitem[Qin et~al.(2024)Qin, Liang, Ye, Zhu, Yan, Lu, Lin, Cong, Tang, Qian,
  et~al.]{qin2024toolllm}
Yujia Qin, Shihao Liang, Yining Ye, Kunlun Zhu, Lan Yan, Yaxi Lu, Yankai Lin,
  Xin Cong, Xiangru Tang, Bill Qian, et~al.
\newblock {ToolLLM}: Facilitating large language models to master 16000+
  real-world {APIs}.
\newblock In \emph{International Conference on Learning Representations},
  volume 2024, pages 9695--9717, 2024.

\bibitem[Qu et~al.(2025)Qu, Dai, Wei, Cai, Wang, Yin, Xu, and Wen]{qu2025tool}
Changle Qu, Sunhao Dai, Xiaochi Wei, Hengyi Cai, Shuaiqiang Wang, Dawei Yin,
  Jun Xu, and Ji-Rong Wen.
\newblock Tool learning with large language models: A survey.
\newblock \emph{Frontiers of Computer Science}, 19\penalty0 (8):\penalty0
  198343, 2025.

\bibitem[Qu et~al.(2026)Qu, Dai, Cai, Xu, Wang, and Yin]{qu2026matchtir}
Changle Qu, Sunhao Dai, Hengyi Cai, Jun Xu, Shuaiqiang Wang, and Dawei Yin.
\newblock {MatchTIR}: Fine-grained supervision for tool-integrated reasoning
  via bipartite matching.
\newblock In \emph{Proceedings of the 64th Annual Meeting of the Association
  for Computational Linguistics (Volume 1: Long Papers)}, pages 11953--11968,
  2026.

\bibitem[Sagawa et~al.(2020)Sagawa, Koh, Hashimoto, and
  Liang]{sagawa2020distributionally}
Shiori Sagawa, Pang~Wei Koh, Tatsunori~B. Hashimoto, and Percy Liang.
\newblock Distributionally robust neural networks.
\newblock In \emph{International Conference on Learning Representations}, 2020.
\newblock URL \url{https://openreview.net/forum?id=ryxGuJrFvS}.

\bibitem[Schick et~al.(2023)Schick, Dwivedi-Yu, Dess{\`\i}, Raileanu, Lomeli,
  Hambro, Zettlemoyer, Cancedda, and Scialom]{schick2023toolformer}
Timo Schick, Jane Dwivedi-Yu, Roberto Dess{\`\i}, Roberta Raileanu, Maria
  Lomeli, Eric Hambro, Luke Zettlemoyer, Nicola Cancedda, and Thomas Scialom.
\newblock {Toolformer}: Language models can teach themselves to use tools.
\newblock \emph{Advances in neural information processing systems},
  36:\penalty0 68539--68551, 2023.

\bibitem[Shao et~al.(2024)Shao, Wang, Zhu, Xu, Song, Bi, Zhang, Zhang, Li, Wu,
  et~al.]{shao2024deepseekmath}
Zhihong Shao, Peiyi Wang, Qihao Zhu, Runxin Xu, Junxiao Song, Xiao Bi, Haowei
  Zhang, Mingchuan Zhang, YK~Li, Yang Wu, et~al.
\newblock {DeepSeekMath}: Pushing the limits of mathematical reasoning in open
  language models.
\newblock \emph{arXiv preprint arXiv:2402.03300}, 2024.

\bibitem[Shen et~al.(2023)Shen, Song, Tan, Li, Lu, and
  Zhuang]{shen2023hugginggpt}
Yongliang Shen, Kaitao Song, Xu~Tan, Dongsheng Li, Weiming Lu, and Yueting
  Zhuang.
\newblock Hugginggpt: Solving ai tasks with chatgpt and its friends in hugging
  face.
\newblock \emph{Advances in Neural Information Processing Systems},
  36:\penalty0 38154--38180, 2023.

\bibitem[Shi et~al.(2024)Shi, Yuan, Wu, Wang, and Feng]{shi2024dmpo}
Wentao Shi, Mengqi Yuan, Junkang Wu, Qifan Wang, and Fuli Feng.
\newblock Direct multi-turn preference optimization for language agents.
\newblock In \emph{Conference on Empirical Methods in Natural Language
  Processing (EMNLP)}, 2024.

\bibitem[Shinn et~al.(2023)Shinn, Cassano, Gopinath, Narasimhan, and
  Yao]{shinn2023reflexion}
Noah Shinn, Federico Cassano, Ashwin Gopinath, Karthik Narasimhan, and Shunyu
  Yao.
\newblock {Reflexion}: Language agents with verbal reinforcement learning.
\newblock \emph{Advances in neural information processing systems},
  36:\penalty0 8634--8652, 2023.

\bibitem[Song et~al.(2024)Song, Yin, Yue, Huang, Li, and Lin]{song2024trial}
Yifan Song, Da~Yin, Xiang Yue, Jie Huang, Sujian Li, and Bill~Yuchen Lin.
\newblock Trial and error: Exploration-based trajectory optimization of {LLM}
  agents.
\newblock In \emph{Proceedings of the 62nd Annual Meeting of the Association
  for Computational Linguistics (Volume 1: Long Papers)}, pages 7584--7600,
  2024.

\bibitem[Su et~al.(2026)Su, Wan, Yang, Shi, Han, Qiu, and Luo]{su2026failure}
Junhao Su, Yuanliang Wan, Junwei Yang, Hengyu Shi, Tianyang Han, Yurui Qiu, and
  Junfeng Luo.
\newblock Failure makes the agent stronger: Enhancing accuracy through
  structured reflection for reliable tool interactions.
\newblock In \emph{Findings of the Association for Computational Linguistics:
  ACL 2026}, pages 12712--12734, 2026.

\bibitem[Sullivan et~al.(2025)Sullivan, Hartmann, and
  Koller]{sullivan2025procedural}
Michael Sullivan, Mareike Hartmann, and Alexander Koller.
\newblock Procedural environment generation for tool-use agents.
\newblock In \emph{Proceedings of the 2025 Conference on Empirical Methods in
  Natural Language Processing}, pages 18555--18573, 2025.

\bibitem[Sun and Saenko(2016)]{sun2016deep}
Baochen Sun and Kate Saenko.
\newblock Deep coral: Correlation alignment for deep domain adaptation.
\newblock In \emph{European conference on computer vision}, pages 443--450.
  Springer, 2016.

\bibitem[Tang et~al.(2023)Tang, Deng, Lin, Han, Liang, Cao, and
  Sun]{tang2023toolalpaca}
Qiaoyu Tang, Ziliang Deng, Hongyu Lin, Xianpei Han, Qiao Liang, Boxi Cao, and
  Le~Sun.
\newblock {ToolAlpaca}: Generalized tool learning for language models with 3000
  simulated cases.
\newblock \emph{arXiv preprint arXiv:2306.05301}, 2023.

\bibitem[Tao et~al.(2026)Tao, Peng, Yao, Ge, Cheng, Wang, Gao, and
  Li]{tao2026trace}
Leitian Tao, Baolin Peng, Wenlin Yao, Tao Ge, Hao Cheng, Mike~Hang Wang,
  Jianfeng Gao, and Sharon Li.
\newblock {TRACE}: Turn-level reward assignment via credit estimation for
  long-horizon agents.
\newblock \emph{arXiv preprint arXiv:2607.13988}, 2026.

\bibitem[Team(2026)]{qwen35blog}
Qwen Team.
\newblock {Qwen3.5}: Accelerating productivity with native multimodal agents,
  February 2026.
\newblock URL \url{https://qwen.ai/blog?id=qwen3.5}.

\bibitem[Trivedi et~al.(2024)Trivedi, Khot, Hartmann, Manku, Dong, Li, Gupta,
  Sabharwal, and Balasubramanian]{trivedi2024appworld}
Harsh Trivedi, Tushar Khot, Mareike Hartmann, Ruskin Manku, Vinty Dong, Edward
  Li, Shashank Gupta, Ashish Sabharwal, and Niranjan Balasubramanian.
\newblock {AppWorld}: A controllable world of apps and people for benchmarking
  interactive coding agents.
\newblock In \emph{Proceedings of the 62nd Annual Meeting of the Association
  for Computational Linguistics (Volume 1: Long Papers)}, pages 16022--16076,
  2024.

\bibitem[Wu et~al.(2026)Wu, Meij, and Yilmaz]{wu2026beyond}
Bin Wu, Edgar Meij, and Emine Yilmaz.
\newblock Beyond static toolsets: Self-evolving llm tool agents via continual
  documentation adaptation.
\newblock In \emph{Findings of the Association for Computational Linguistics:
  ACL 2026}, pages 21519--21539, 2026.

\bibitem[Wu et~al.(2025)Wu, Zhu, Han, Zhang, Shao, and Chen]{wu2025chain}
Mengsong Wu, Tong Zhu, Han Han, Xiang Zhang, Wenbiao Shao, and Wenliang Chen.
\newblock Chain-of-tools: Utilizing massive unseen tools in the cot reasoning
  of frozen language models.
\newblock \emph{arXiv preprint arXiv:2503.16779}, 2025.

\bibitem[Xi et~al.(2026)Xi, Guo, Liu, Zhang, Fan, Zhang, Liu, Chai, Shi, Zhai,
  et~al.]{xi2026can}
Zhiheng Xi, Xin Guo, Jiaqi Liu, Jiazheng Zhang, Yutao Fan, Zhihao Zhang,
  Shichun Liu, Mingxu Chai, Xiaowei Shi, Yitao Zhai, et~al.
\newblock Can rl improve generalization of llm agents? an empirical study.
\newblock \emph{arXiv preprint arXiv:2603.12011}, 2026.

\bibitem[Xie et~al.(2026)Xie, Thomas, Hansen, Fu, Li, and Wang]{xie2026tips}
Yutao Xie, Nathaniel Thomas, Nick Hansen, Yang Fu, Li~Li, and Xiaolong Wang.
\newblock {TIPS}: Turn-level information-potential reward shaping for
  search-augmented {LLMs}.
\newblock In \emph{International Conference on Learning Representations},
  volume 2026, pages 156549--156584, 2026.

\bibitem[Xu et~al.(2026)Xu, Lin, Xue, Wang, Xu, Wu, Zhang, Lin, Dong, Ling,
  et~al.]{xu2026deepseek}
Anyi Xu, Bangcai Lin, Bing Xue, Bingxuan Wang, Bingzheng Xu, Bochao Wu, Bowei
  Zhang, Chaofan Lin, Chen Dong, Chenchen Ling, et~al.
\newblock Deepseek-v4: Towards highly efficient million-token context
  intelligence.
\newblock \emph{arXiv preprint arXiv:2606.19348}, 2026.

\bibitem[Xue et~al.(2026)Xue, Zheng, Liu, Li, Zheng, Ma, and
  An]{xue2026simpletir}
Zhenghai Xue, Longtao Zheng, Qian Liu, Yingru Li, Xiaosen Zheng, Zejun Ma, and
  Bo~An.
\newblock Simpletir: End-to-end reinforcement learning for multi-turn
  tool-integrated reasoning.
\newblock In \emph{International Conference on Learning Representations},
  volume 2026, pages 8424--8449, 2026.

\bibitem[Yao et~al.(2023)Yao, Zhao, Yu, Du, Shafran, Narasimhan, and
  Cao]{yao2023react}
Shunyu Yao, Jeffrey Zhao, Dian Yu, Nan Du, Izhak Shafran, Karthik~R Narasimhan,
  and Yuan Cao.
\newblock {ReAct}: Synergizing reasoning and acting in language models.
\newblock In \emph{The Eleventh International Conference on Learning
  Representations}, 2023.

\bibitem[Ye et~al.(2026)Ye, Jiang, Du, Xu, Yao, Xi, Fan, Zhang, Gui, Huang,
  et~al.]{ye2026feedback}
Junjie Ye, Changhao Jiang, Zhengyin Du, Yufei Xu, Xuesong Yao, Zhiheng Xi,
  Xiaoran Fan, Qi~Zhang, Tao Gui, Xuan-Jing Huang, et~al.
\newblock Feedback-driven tool-use improvements in large language models via
  automated build environments.
\newblock In \emph{Findings of the Association for Computational Linguistics:
  ACL 2026}, pages 2293--2323, 2026.

\bibitem[Yu et~al.(2025)Yu, Wang, Ma, Wang, Wu, Guo, and Zhang]{yu2025steptool}
Yuanqing Yu, Zhefan Wang, Weizhi Ma, Shuai Wang, Chuhan Wu, Zhiqiang Guo, and
  Min Zhang.
\newblock {StepTool}: Enhancing multi-step tool usage in {LLMs} via
  step-grained reinforcement learning.
\newblock In \emph{Proceedings of the 34th ACM International Conference on
  Information and Knowledge Management}, pages 3952--3962, 2025.

\bibitem[Yuan et~al.(2023)Yuan, Yuan, Li, Dong, Lu, Tan, Zhou, and
  Zhou]{yuan2023rft}
Zheng Yuan, Hongyi Yuan, Chengpeng Li, Guanting Dong, Keming Lu, Chuanqi Tan,
  Chang Zhou, and Jingren Zhou.
\newblock Scaling relationship on learning mathematical reasoning with large
  language models.
\newblock \emph{arXiv preprint arXiv:2308.01825}, 2023.

\bibitem[Zeng et~al.(2026)Zeng, Lv, Hou, Du, Zheng, Chen, Yin, Ge, Huang, Xie,
  et~al.]{zeng2026glm}
Aohan Zeng, Xin Lv, Zhenyu Hou, Zhengxiao Du, Qinkai Zheng, Bin Chen, Da~Yin,
  Chendi Ge, Chenghua Huang, Chengxing Xie, et~al.
\newblock Glm-5: from vibe coding to agentic engineering.
\newblock \emph{arXiv preprint arXiv:2602.15763}, 2026.

\bibitem[Zeng et~al.(2025{\natexlab{a}})Zeng, Wei, Brown, Frunza, Nevmyvaka,
  Zhao, and Hong]{zeng2025reinforcing}
Siliang Zeng, Quan Wei, William Brown, Oana Frunza, Yuriy Nevmyvaka, Yang~Katie
  Zhao, and Mingyi Hong.
\newblock Reinforcing multi-turn reasoning in {LLM} agents via turn-level
  credit assignment.
\newblock In \emph{ICML 2025 Workshop on Computer Use Agents},
  2025{\natexlab{a}}.

\bibitem[Zeng et~al.(2025{\natexlab{b}})Zeng, Ding, Hou, Wang, Du, Dai, Ding,
  Tang, Tu, Liu, et~al.]{zeng2025tool}
Yirong Zeng, Xiao Ding, Yutai Hou, Yuxian Wang, Li~Du, Juyi Dai, Qiuyang Ding,
  Duyu Tang, Dandan Tu, Weiwen Liu, et~al.
\newblock Tool zero: Training tool-augmented llms via pure rl from scratch.
\newblock In \emph{EMNLP (Findings)}, pages 9135--9147, 2025{\natexlab{b}}.

\bibitem[Zheng et~al.(2023)Zheng, Chiang, Sheng, Zhuang, Wu, Zhuang, Lin, Li,
  Li, Xing, et~al.]{zheng2023judging}
Lianmin Zheng, Wei-Lin Chiang, Ying Sheng, Siyuan Zhuang, Zhanghao Wu, Yonghao
  Zhuang, Zi~Lin, Zhuohan Li, Dacheng Li, Eric Xing, et~al.
\newblock Judging llm-as-a-judge with mt-bench and chatbot arena.
\newblock \emph{Advances in neural information processing systems},
  36:\penalty0 46595--46623, 2023.

\bibitem[Zhuang et~al.(2026)Zhuang, Zhang, Zhou, Zhang, and
  Huang]{zhuang2026losemb}
Luyao Zhuang, Qinggang Zhang, Huachi Zhou, Yujing Zhang, and Xiao Huang.
\newblock Losemb: Logic-guided semantic bridging for inductive tool retrieval.
\newblock In \emph{Proceedings of the ACM Web Conference 2026}, pages
  3835--3846, 2026.

\end{thebibliography}
\endgroup
\startappendices
\raggedbottom

\section{Additional Experimental Details}
\label{app}
\label{app:experimental-details}

\subsection{Dataset Details}
\label{app:dataset-details}

We evaluate \method on two multi-turn tool-use benchmarks, AppWorld and FTRL, both of which provide executable tools and verifiable task outcomes. For each benchmark, we describe the dataset contents, the ID/OOD split construction, and the evaluation protocol below.

\textbf{AppWorld.}
We use the publicly released AppWorld benchmark~\citep{trivedi2024appworld},\footnote{\url{https://github.com/StonyBrookNLP/appworld}} with 90 training tasks across 30 scenarios and 473 APIs. The benchmark provides two official test splits: Test-C contains all tasks that require at least one API from the designated unseen applications (i.e., Amazon and Gmail), and Test-N contains the test tasks that do not require them. Therefore, we treat Test-N as the ID test set and Test-C as the OOD test set, containing 168 and 417 tasks, respectively. No training task requires the unseen applications. Each task is scored by the official state-based evaluator, which runs task-specific tests against the final environment state. Post-training uses the outcome reward returned by this evaluator. We report the task success rate, i.e., the percentage of tasks that pass all evaluation tests. Results under the stricter Scenario Goal Completion (SGC) metric are provided in Appendix~\ref{app:sgc}.

\textbf{FTRL.}
FTRL~\citep{ye2026feedback} covers single-hop and multi-hop tool-use queries, each associated with a subject domain. To construct a covariate-shift OOD split, we hold out five domains: Health, Medicine and Public Health; Music and Performing Arts; Politics, Governance and Law; Sports and Competitions; and Zoology and Animal Science. Their tools are excluded during post-training, while their function classes remain covered by the seen domains. The test set contains 168 ID queries and 32 OOD queries; an OOD query requires at least one held-out-domain tool. Post-training uses 1,615 training instances that do not require these tools. Following the official protocol, we compute the metrics per query: for a trajectory with \(p\) tool calls that solves \(q\) of the \(n\) required sub-questions, Solve-P \(=q/p\) measures tool-invocation precision, Solve-R \(=q/n\) measures task-completion recall, and Solve-F1 is their harmonic mean.

\subsection{Input Prompts}
\label{app:input-prompts}

We use the official task prompts of AppWorld~\citep{trivedi2024appworld} and FTRL~\citep{ye2026feedback}, shown in Figures~\ref{fig:appworld-react-prompt} and~\ref{fig:ftrl-react-prompt}.

\par
\begin{figure}[t]
\begin{promptbox}[unbreakable]{AppWorld ReAct Prompt}
I am your supervisor and you are a super intelligent AI Assistant whose job is
to achieve my day-to-day tasks completely autonomously.

You will interact with apps using their associated APIs through a multi-step
conversation in a Python REPL. Write Python code; the environment will execute
it and return the result, which you can use in the next step.

# List the available apps.
print(apis.api_docs.show_app_descriptions())

# List the APIs of an app.
print(apis.api_docs.show_api_descriptions(app_name='<app_name>'))

# Show the specification of an API.
print(apis.api_docs.show_api_doc(
    app_name='<app_name>', api_name='<api_name>'))

Use only the provided APIs. Write one small code block at each step and use
results from previous steps when needed. When the task is complete, call
apis.supervisor.complete_task(); if an answer is required, pass it through the
answer argument.

My name is: {supervisor_first_name} {supervisor_last_name}.
My personal email is {supervisor_email} and phone number is
{supervisor_phone_number}.

Your task is: {task_description}

Reason about the next API call within <think> </think> tags and place the code
body within <code> </code> tags.
\end{promptbox}

\promptcaption{Task-facing AppWorld ReAct prompt. The demonstration and standard environment disclaimers in the released prompt are omitted for space.}{fig:appworld-react-prompt}
\end{figure}

\par

\par
\begin{figure}[t]
\begin{promptbox}[unbreakable]{FTRL ReAct Prompt}
Please call given tools to answer the question. Please note that all your
information must be obtained by calling tools and not by answering the question
directly. If the call fails, you need to try to correct it and continue until
you arrive at an answer.

Question: {question}
\end{promptbox}

\promptcaption{Task prompt used for FTRL. The query-specific tool schemas are supplied through the native tool-calling interface.}{fig:ftrl-react-prompt}
\end{figure}

\par

\subsection{Implementation Details}
\label{app:implementation-details}

\textbf{General setup.}
We use Qwen3.5-4B and Qwen3.5-9B~\citep{qwen35blog} as policy backbones and adopt the same ReAct interaction scaffold~\citep{yao2023react} for all methods. Unless otherwise specified, \method uses GRPO as the host objective. All trainable methods are optimized with Adam. Following AppWorld~\citep{trivedi2024appworld}, we select hyperparameters on its official development split. On FTRL~\citep{ye2026feedback}, we follow MatchTIR's released protocol~\citep{qu2026matchtir}, using the test split for hyperparameter selection. We apply the same selection protocol to \method and all baselines. The default online training configurations for both benchmarks are summarized in Table~\ref{tab:training-configuration}.

\textbf{Interaction and inference.}
We follow the released interaction protocol of each benchmark. In AppWorld, the agent retrieves app descriptions and API specifications on demand through the \texttt{api\_docs} interface, and the specifications of Amazon and Gmail become available only at evaluation time. In FTRL, each query is paired with a fixed candidate set of tool schemas, which is passed to the policy through the native tool-calling interface. Training rollouts and evaluation both use sampled decoding with temperature 1.0 and top-\(p=1.0\). Each reported three-seed result trains and evaluates the complete pipeline with three random seeds, which control the initialization of the projection head and prototypes, the data order, and rollout sampling.

\textbf{\method configuration.}
For each tool call, we extract the hidden state of the call's final token from layer \(l=8\) of the policy backbone; Appendix~\ref{app:representation-extraction} illustrates the extraction. The projection head \(g_\psi\) is implemented as a two-layer MLP with output dimension \(d=128\). Function prototypes are randomly initialized as unit vectors and updated with an exponential moving average (EMA). We set the temperature \(\tau=0.4\), the loss weights \(\lambda_{\mathrm{vMF}}=0.02\) and \(\lambda_{\mathrm{sep}}=2\), and the EMA factor \(\alpha=0.9\). Both the projection head and the prototypes are not used at deployment, so \method adds no inference-time overhead. For the distance-metric ablation in Figure~\ref{fig:design-ablation}(b), we replace the cosine score $\mathbf{a}^{\top}\mathbf{b}/\tau$ with the negative unsquared Euclidean score $-\lVert\mathbf{a}-\mathbf{b}\rVert_2/\tau$ in both $\mathcal{L}_{\mathrm{var}}$ and $\mathcal{L}_{\mathrm{sep}}$. Here, $\mathbf{a}$ is the unit-normalized tool-call representation or batch direction, and $\mathbf{b}$ is the unit prototype. Both variants use $\tau=0.4$, with the projection head, EMA update, and loss weights unchanged.

\begin{table}[t]
    \centering
    \caption{Default online training configuration.}
    \label{tab:training-configuration}
\normalsize
    \setlength{\tabcolsep}{7pt}
    \renewcommand{\arraystretch}{1.05}
    \begin{tabular}{@{}lcc@{}}
        \toprule
        Hyperparameter & AppWorld & FTRL \\
        \midrule
        Prompts per update & 40 & 256 \\
        Rollouts per prompt & 6 & 16 \\
        Global batch size & 240 & 4,096 \\
        Maximum training turns & 40 & 10 \\
        Maximum evaluation turns & 40 & 20 \\
        Maximum generated tokens per turn & 2,000 & 4,096 \\
        Maximum response length & 2,000 & 23,000 \\
        Learning rate & \multicolumn{2}{c}{\(1\times10^{-6}\)} \\
        Weight decay & \multicolumn{2}{c}{0.01} \\
        KL coefficient & \multicolumn{2}{c}{0.001} \\
        Entropy coefficient & \multicolumn{2}{c}{0.001} \\
        Clipping range & \multicolumn{2}{c}{0.2} \\
        Gradient clipping & \multicolumn{2}{c}{1.0} \\
        Training duration & 10 epochs & 3 epochs \\
        \bottomrule
    \end{tabular}
    
\end{table}

\subsection{Tool-Call Representation Extraction}
\label{app:representation-extraction}

\method represents each tool call by the hidden state of the call's final token. When a code block contains multiple API calls, we extract a separate representation for each call at its own final token. Figure~\ref{fig:representation-extraction} illustrates a code block with two API calls. For call $i$, we take the layer-\(l=8\) hidden state at its final token as \(\mathbf{h}_i\) and compute \(\mathbf{z}_i=\operatorname{Normalize}(g_\psi(\mathbf{h}_i))\). The two calls therefore contribute two representations to the representation-shaping objective. The same extraction rule applies to the FTRL tool-calling format.

\par
\begin{figure}[t]
\begin{promptbox}[unbreakable]{Tool-Call Representation Extraction}
<think> I first find the artist, then search for their songs. </think>
<code>
artists = apis.spotify.search_artists(query="Lily Moon")  # call 1 ends here
artist_id = artists[0]["artist_id"]
songs = apis.spotify.search_songs(
    query="Lily Moon", artist_id=artist_id
)  # call 2 ends here
</code>
\end{promptbox}

\promptcaption{Tool-call representation extraction from a code block containing two AppWorld API calls. The final token of each call provides its own layer-\(l=8\) hidden state, \(\mathbf{h}_1\) and \(\mathbf{h}_2\), which are independently projected and normalized.}{fig:representation-extraction}
\end{figure}

\par

\subsection{Domain and Function-Class Annotations}
\label{app:function-annotations}

Following \citet{zheng2023judging}, we use GPT-5.6-sol to annotate the tools in both benchmarks with domain labels and function-class labels.

\textbf{Domain annotation.}
Both benchmarks share the same domain-annotation prompt, shown in Figure~\ref{fig:domain-prompt}; the closed taxonomy is instantiated per benchmark. This yields nine application domains for AppWorld and 27 subject domains for FTRL. The ten AppWorld meta APIs (\texttt{api\_docs} and \texttt{supervisor}) are infrastructure and receive no domain label.

\par
\begin{figure}[t]
\begin{promptbox}[unbreakable]{Domain Annotation Prompt}
You label software tools by their DOMAIN.

The domain is the real-world topic or application area that a tool
operates on, not the operation it performs. Use the tool description as the primary
evidence. The tool name is supporting evidence only. Choose exactly one
domain from the closed taxonomy below:

{domain_taxonomy}

Important boundaries:
- Judge the topic of the data the tool reads or writes, not the action
  type; functional intent is labeled separately.
- A tool that searches, computes, or verifies within one topic area
  belongs to that area.
- If several domains are involved, label the dominant one. Lower
  confidence when the description is genuinely underspecified.

Input:
tool_name: {tool_name}
description: {description}

Return strict JSON only:
{"domain": "<ONE_DOMAIN>", "reason": "<one sentence>",
 "confidence": <0.0--1.0>}
\end{promptbox}

\promptcaption{Prompt used to annotate tool domains with GPT-5.6-sol. The closed taxonomy is instantiated per benchmark.}{fig:domain-prompt}
\end{figure}

\par

\textbf{Function-class annotation.}
GPT-5.6-sol assigns a function-class label to each of the 473 AppWorld APIs and 4,545 FTRL tools. The 463 ordinary AppWorld APIs are grouped into 14 function classes, while the ten meta APIs are labeled META and excluded from the representation-shaping objective; the FTRL tools are grouped into nine function classes. Only the labels of tools observed during post-training enter the objective, and the annotations of OOD tools are never accessed. The prompts are shown in Figures~\ref{fig:appworld-function-prompt} and~\ref{fig:ftrl-function-prompt}, and Table~\ref{tab:function-class-statistics} summarizes the statistics of the resulting annotations.

\par
\begin{figure}[t]
\begin{promptbox}[unbreakable,listing options={style=radio-prompt,basicstyle=\ttfamily\fontsize{9}{12.5}\selectfont}]{AppWorld Annotation Prompt}
You label AppWorld APIs by their dominant externally visible function. Classify
what the API does, rather than its application, domain, or surface name. Choose
exactly one family from the closed taxonomy below:

- AUTH_ACTION: Manage an account, authentication, authorization, or session
  lifecycle.
- CHECK_STATE: Check existence or make an explicit state judgment.
- CONTROL_ACTION: Control a transient process, such as playback or a device.
- CREATE_OBJECT: Create a new record, object, or piece of content.
- DELETE_ACTION: Delete, cancel, or remove an object.
- ORGANIZE_ACTION: Change membership, order, location, or association.
- READ_ITEM: Retrieve a known item, detail, scalar value, or state.
- READ_LIST: Search, discover, filter, or enumerate candidate items.
- SEND_ACTION: Send, reply, forward, invite, notify, or remind.
- TOGGLE_ACTION: Change a discrete flag or lifecycle status.
- TRANSFER_ACTION: Move money or assets between parties.
- TRANSFORM_ACTION: Copy, compress, decompress, convert, or transform supplied
  content.
- UPDATE_OBJECT: Modify fields or content of an existing object.
- WORKFLOW_ACTION: Commit an order, return, subscription, approval, or related
  application workflow.

Important boundaries:
- READ_LIST returns a collection; READ_ITEM returns one known item or property.
- UPDATE_OBJECT changes free-form content; TOGGLE_ACTION changes a discrete
  status; CONTROL_ACTION controls a transient process.
- ORGANIZE_ACTION changes a relationship or collection without creating the
  primary object.
- Use TRANSFER_ACTION whenever money or assets move, even if the API name
  contains "create" or "record."
- Reserve WORKFLOW_ACTION for application-level commitments.
- If the API belongs to api_docs or supervisor, return META.META; otherwise
  choose one of the 14 families above.

Input:
tool_name: {tool_name}
description: {description}
arguments: {arguments}
response: {response_specification}

Return strict JSON only:
{"family": "<ONE_FAMILY>", "reason": "<one sentence>",
 "confidence": <0.0--1.0>}
\end{promptbox}

\promptcaption{Prompt used to annotate AppWorld APIs with GPT-5.6-sol.}{fig:appworld-function-prompt}
\end{figure}

\par

\par
\begin{figure}[t]
\begin{promptbox}[unbreakable,listing options={style=radio-prompt,basicstyle=\ttfamily\fontsize{9}{12.5}\selectfont}]{FTRL Annotation Prompt}
You label software tools by their dominant FUNCTIONAL INTENT.

Use the tool description as the primary evidence. The tool name is supporting
evidence only. Choose exactly one family from the closed taxonomy below:

- RETRIEVE_PROVIDE: Retrieve or provide factual information about a known
  subject, item, record, or property; no discovery, comparison, verification,
  or transformation is dominant.
- SEARCH_LOCATE: Find, search, query, filter, or locate candidate entities,
  people, places, records, works, or resources.
- IDENTIFY_CLASSIFY: Determine identity, type, category, author,
  creator, owner, origin, membership, or another categorical attribute.
- ANALYZE_SUMMARIZE: Analyze, explain, synthesize, interpret, correlate, or
  summarize information rather than merely retrieving one fact.
- CALCULATE_MEASURE: Calculate, count, estimate, aggregate, convert, measure,
  or derive a numeric value.
- COMPARE_SELECT: Compare, rank, score, optimize, recommend, or select among
  alternatives, including superlatives.
- TRACK_FORECAST: Track or monitor change over time, reconstruct progression or
  history, or forecast a future value or state.
- VERIFY_VALIDATE: Check, verify, validate, test, assess compliance,
  availability, or consistency, or return a yes/no validity judgment.
- TRANSFORM_EXTRACT: Extract structured content, parse, translate, normalize,
  convert representation, map, or transform supplied content.

Important boundaries:
- SEARCH_LOCATE discovers candidate entities; RETRIEVE_PROVIDE fetches facts
  about a subject already specified by the caller.
- IDENTIFY_CLASSIFY resolves identity, type, category, origin, or ownership.
- COMPARE_SELECT includes ranking, recommendation, best, nearest, largest, and
  choosing among candidates.
- VERIFY_VALIDATE is an explicit check or validity judgment, not an ordinary
  lookup.
- CALCULATE_MEASURE must have a derived numeric result as the dominant
  function.
- TRACK_FORECAST concerns temporal progression or prediction.
- TRANSFORM_EXTRACT operates on supplied content or changes its representation.
- If several operations are mentioned, label the dominant externally visible
  result. Lower confidence when the description is genuinely underspecified.

Input:
tool_name: {tool_name}
description: {description}

Return strict JSON only:
{"family": "<ONE_FAMILY>", "primitive": "<concise verb phrase>",
 "reason": "<one sentence>", "confidence": <0.0--1.0>}
\end{promptbox}

\promptcaption{Prompt used to annotate FTRL tools with GPT-5.6-sol.}{fig:ftrl-function-prompt}
\end{figure}

\par

\textbf{Annotation verification.}
To assess annotation quality, three annotators independently verify all 463 AppWorld APIs and a random sample of 200 FTRL tools, and we compute Fleiss's kappa over the three sets of labels. The kappa scores are 0.86 on AppWorld and 0.83 on FTRL, with raw agreement rates of 88.77\% and 85.17\%, respectively; the remaining disagreements are resolved by discussion. The strong agreement indicates that the function classes can be consistently identified from tool specifications.

\begin{table}[t]
    \centering
    \caption{Statistics of the function-class annotations. The ten AppWorld meta APIs are excluded from the representation-shaping objective.}
    \label{tab:function-class-statistics}
\normalsize
    \setlength{\tabcolsep}{6pt}
    \renewcommand{\arraystretch}{1.03}
    \begin{tabular}{@{}lrrr@{}}
        \toprule
        Dataset & Tools/APIs & Function classes & Average class size \\
        \midrule
        AppWorld & 463 & 14 & 33.1 \\
        FTRL & 4,545 & 9 & 505.0 \\
        \bottomrule
    \end{tabular}
    
\end{table}

\subsection{Baseline Details}
\label{app:baselines}

We follow the baseline formulations in the cited papers and use the interaction and evaluation protocols described in Appendix~\ref{app:implementation-details}. Baseline-specific objectives are summarized below.

\textbf{Prompting-based agents.}
GPT-5.5~\citep{openai2026gpt55}, Claude Opus 4.8~\citep{anthropic2026claudeopus48}, GLM-5.2~\citep{zeng2026glm}, and DeepSeek V4 Pro~\citep{xu2026deepseek} are evaluated without any post-training. Each model receives the same ReAct prompt, tool specifications, observations, and interaction budget as in the main setup.

\textbf{General post-training objectives.}
GRPO~\citep{shao2024deepseekmath} optimizes the outcome reward and applies the same trajectory-level advantage to all tool calls in a rollout. RFT~\citep{yuan2023rft} fine-tunes on successful rollouts selected by the task evaluator, while DMPO~\citep{shi2024dmpo} constructs trajectory-level preference pairs from the same evaluator scores.

\textbf{Turn-level tool-use post-training methods.}
StepTool~\citep{yu2025steptool} and FTRL-M~\citep{ye2026feedback} assign turn-level rewards, where FTRL-M denotes the multi-turn variant of FTRL. MatchTIR~\citep{qu2026matchtir} uses the KM variant with AppWorld's reference API calls and FTRL's released tool-call annotations, combining turn-level and trajectory-level advantages. SOAR~\citep{li2026soar} derives supervision from observation tokens, while TRACE~\citep{tao2026trace} follows the authors' turn-level credit-assignment setup.

\textbf{Tool-use RL methods.}
SimpleTIR~\citep{xue2026simpletir} applies void-turn filtering during training, and ToolMaster~\citep{gao2026teaching} combines trajectory-based supervised fine-tuning with subsequent reinforcement learning. LOOP~\citep{chen2025reinforcement} retains its leave-one-out advantage estimation and rollout reuse.

\textbf{Representation-learning method.}
SEAL~\citep{li2026cyclical} is run with its released implementation without further modification.

\textbf{OOD-generalization methods.}
CORAL~\citep{sun2016deep} aligns representation covariances, while Group DRO~\citep{sagawa2020distributionally} reweights training-group losses. On AppWorld, CORAL aligns tool-call representations across training applications, and Group DRO reweights application-group policy losses. ToolRL~\citep{qian2026toolrl} and PAFT~\citep{lv2026can} use their tool-call reward and training-time trajectory perturbations, respectively.

\subsection{Compute Resources and Time}
\label{app:compute}

\textbf{Software and hardware.}
We conduct all experiments using Python 3.12.13 and PyTorch 2.11.0 with CUDA 12.8 on four NVIDIA H200 GPUs with 141\,GB memory.

\textbf{Training time.}
With Qwen3.5-4B, a complete \method run takes approximately 6.6 hours on AppWorld and 21.9 hours on FTRL. Compared with vanilla GRPO under the same configuration, the projection head, the prototype updates, and the representation-shaping losses add less than 1\% training time.

\clearpage
\begin{algorithm}[t]
\caption{Post-training with \method}
\label{alg:toolcompass}
\label{alg}

\KwInput{Policy $\pi_\theta$; projection head $g_\psi$; function labels $c(u)$; host objective $\mathcal{L}_{\mathrm{post}}$; layer $l$; temperature $\tau$; loss weights $\lambda_{\mathrm{vMF}}$ and $\lambda_{\mathrm{sep}}$; EMA factor $\alpha$}
Randomly initialize the projection head $g_\psi$ and unit prototypes $\{\boldsymbol{\mu}_k\}_{k\in\mathcal{C}}$\;
\For{each post-training update}{
    Sample a batch of trajectories with $\pi_\theta$ and compute $\mathcal{L}_{\mathrm{post}}$\;
    Initialize the minibatch of tool-call representations $\mathcal{B}\gets\varnothing$\;
    \For{each observed tool call $a_i$ invoking tool $u_i$}{
        Extract $\mathbf{h}_i$ from layer $l$ (Eq.~\ref{eq:hidden}) and compute $\mathbf{z}_i=\operatorname{Normalize}(g_\psi(\mathbf{h}_i))$\;
        Assign $c_i=c(u_i)$ and add $(\mathbf{z}_i,c_i)$ to $\mathcal{B}$\;
}
    Compute $\mathcal{L}_{\mathrm{var}}$ (Eq.~\ref{eq:var}) and $\mathcal{L}_{\mathrm{sep}}$ (Eq.~\ref{eq:sep}) from $\mathcal{B}$, treating the prototypes as stop-gradient targets\;
    Update $\theta$ and $\psi$ using $\mathcal{L}_{\mathrm{post}}+\lambda_{\mathrm{vMF}}(\mathcal{L}_{\mathrm{var}}+\lambda_{\mathrm{sep}}\mathcal{L}_{\mathrm{sep}})$ (Eqs.~\ref{eq:vmf-objective} and~\ref{eq:joint})\;
    \For{each function class $k$ observed in $\mathcal{B}$}{
        Compute its normalized batch direction $\bar{\mathbf{z}}_k$\;
        Update $\boldsymbol{\mu}_k\gets\operatorname{Normalize}(\alpha\boldsymbol{\mu}_k+(1-\alpha)\bar{\mathbf{z}}_k)$ (Eq.~\ref{eq:ema})\;
}
}
\Return{the post-trained policy $\pi_\theta$}\;

\end{algorithm}

\subsection{Training Algorithm}
\label{app:algorithm}

We provide the complete post-training procedure of \method in Algorithm~\ref{alg:toolcompass}, which summarizes the joint optimization of the host objective and the representation-shaping objective, together with the EMA prototype updates.

\section{Additional Analysis}
\label{app:additional-analysis}

Unless otherwise specified, we use Qwen3.5-4B with GRPO as the host objective.

\subsection{Semantic OOD Analysis}
\label{app:semantic-ood}

We evaluate semantic OOD transfer on FTRL by jointly holding out \texttt{CALCULATE\_MEASURE}, \texttt{ANALYZE\_SUMMARIZE}, and \texttt{IDENTIFY\_CLASSIFY} during post-training. GRPO and \method use the same Qwen3.5-4B initialization and filtered training data. We hold out 300 examples from the training data as a validation set. Evaluation includes test queries requiring at least one held-out function. Table~\ref{tab:ftrl-semantic} reports Solve-P, Solve-R, and Solve-F1; other configurations follow Appendix~\ref{app:implementation-details}.

\begin{table}[t]
    \centering
    \caption{Semantic OOD results (\%) on FTRL using Qwen3.5-4B.}
    \label{tab:ftrl-semantic}
\normalsize
    \setlength{\tabcolsep}{12pt}
    \renewcommand{\arraystretch}{1.12}
    \begin{tabular}{@{}lccc@{}}
        \toprule
        Method & Solve-P & Solve-R & Solve-F1 \\
        \midrule
        GRPO & 50.84 & 64.09 & 56.70 \\
        \textbf{\method} & \textbf{53.02} & \textbf{70.21} & \textbf{60.42} \\
        \bottomrule
    \end{tabular}
\end{table}
\suppressfloats[t]

\begin{figure}[t]
    \centering
    \includegraphics[width=0.86\linewidth]{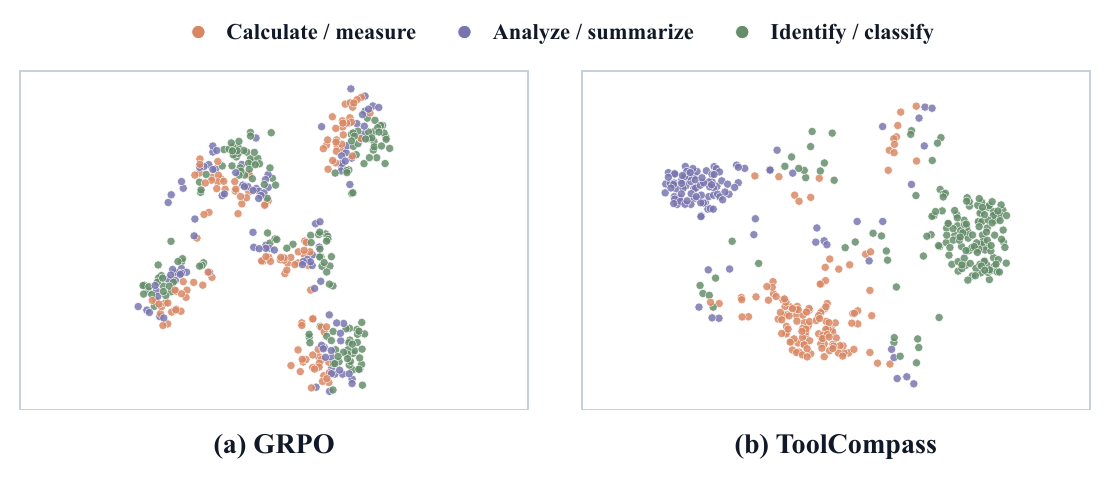}
    
    \caption{ \textbf{t-SNE visualization of jointly held-out functions.}  Colors denote the three function classes excluded from post-training. The two panels compare tool-call representations under GRPO and TOOLCOMPASS.}
    \label{fig:ftrl-semantic-tsne}
    
\end{figure}

Table~\ref{tab:ftrl-semantic} shows that \method improves Solve-P and Solve-R over GRPO by 2.18 and 6.12 points, respectively, on semantic OOD tasks. We attribute this gain to function-level supervision, which encourages the shared backbone to capture operational intent across tools. Together with pretrained semantic knowledge and tool descriptions, this abstraction helps the policy select appropriate calls for functions unseen during post-training.

To further examine this transfer, we visualize tool-call representations for the three held-out functions with t-SNE (Figure~\ref{fig:ftrl-semantic-tsne}). The visualization shows more coherent within-function groups and clearer separation under \method than GRPO, illustrating how function-level organization can extend to unseen functions.

\begin{table}[t]
    
    \centering
    \caption{Comparison of auxiliary objectives on AppWorld with Qwen3.5-4B and GRPO. Values are task success rates (\%).}
    \label{tab:auxiliary-objectives}
\normalsize
    \setlength{\tabcolsep}{13pt}
    \renewcommand{\arraystretch}{1.08}
    \begin{tabular}{@{}lccc@{}}
        \toprule
        Method & ID & OOD & Total \\
        \midrule
        GRPO & 58.33 & 56.83 & 57.26 \\
        GRPO + Function CE & 69.05 & 57.31 & 60.68 \\
        GRPO + SupCon & 60.71 & 61.63 & 61.37 \\
        \textbf{\method} & \textbf{70.63} & \textbf{64.75} & \textbf{66.44} \\
        \bottomrule
    \end{tabular}
    
\end{table}

\subsection{Comparison of Auxiliary Objectives}
\label{app:auxiliary-objectives}

We compare \method with Function CE (linear classification) and SupCon~\citep{khosla2020supervised} under the same function labels, representation layer, projection head, normalization, and training budget. Each objective uses the same development-set tuning budget.

\method outperforms Function CE and SupCon by 7.44 and 3.12 percentage points in OOD task success, respectively, supporting the benefit of the proposed objective beyond generic function supervision.

\begin{figure}[t]
    
    \centering
    \includegraphics[width=0.8\linewidth]{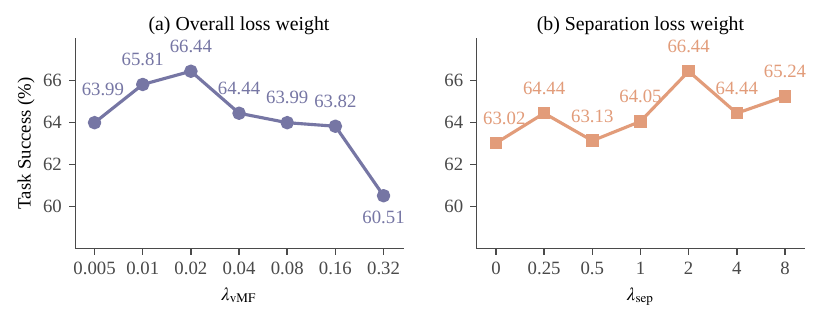}
    
    \caption{ \textbf{Effect of loss weights on AppWorld task success.} We vary one weight while fixing the other to its default value.}
    \label{fig:loss-weights}
    
\end{figure}

\subsection{Effect of Loss Weights}
\label{app:loss-weights}

\method uses two weighting hyperparameters: \(\lambda_{\mathrm{vMF}}\) balances the representation-shaping objective against the host objective, and \(\lambda_{\mathrm{sep}}\) controls the strength of inter-function separation within it. We independently vary each weight while fixing the other to its default value, and report the task success rate in Figure~\ref{fig:loss-weights}.

As shown in Figure~\ref{fig:loss-weights}(a), performance peaks at \(\lambda_{\mathrm{vMF}}=0.02\), reaching 66.44\%, and remains competitive for nearby values, indicating that \method is not overly sensitive to the precise choice of this weight. A very small weight provides insufficient function-level supervision, whereas an overly large weight overemphasizes representation shaping relative to the host objective. Figure~\ref{fig:loss-weights}(b) shows a similar trend for \(\lambda_{\mathrm{sep}}\), which performs best at \(2.0\) with 66.44\%: weaker separation does not sufficiently distinguish different function classes, while stronger separation reduces the flexibility within each class.

\par

\begin{figure}[t]
        \centering
        \includegraphics[width=.55\linewidth]{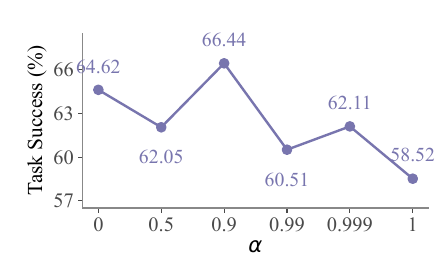}
        
        \caption{ \textbf{Effect of the EMA factor $\alpha$ on AppWorld task success.}}
        \label{fig:ema-ablation}
    \end{figure}

\subsection{Effect of the EMA Factor}
\label{app:ema-ablation}

We vary the EMA factor \(\alpha\) of the prototype update in Eq.~\ref{eq:ema} while keeping all other hyperparameters at their default values. The endpoint \(\alpha=0\) reduces to the batch-estimated prototypes in Figure~\ref{fig:design-ablation}(a), and \(\alpha=1.0\) freezes the prototypes at their initial estimates. As shown in Figure~\ref{fig:ema-ablation}, performance peaks at \(\alpha=0.9\), reaching 66.44\%: a small \(\alpha\) lets the prototypes fluctuate with individual minibatches, while a large \(\alpha\) makes them adapt too slowly to the evolving representations.

\par

\subsection{Qualitative Analysis}
\label{app:qualitative}

\begin{figure}[t]
\centering
\includegraphics[width=\linewidth]{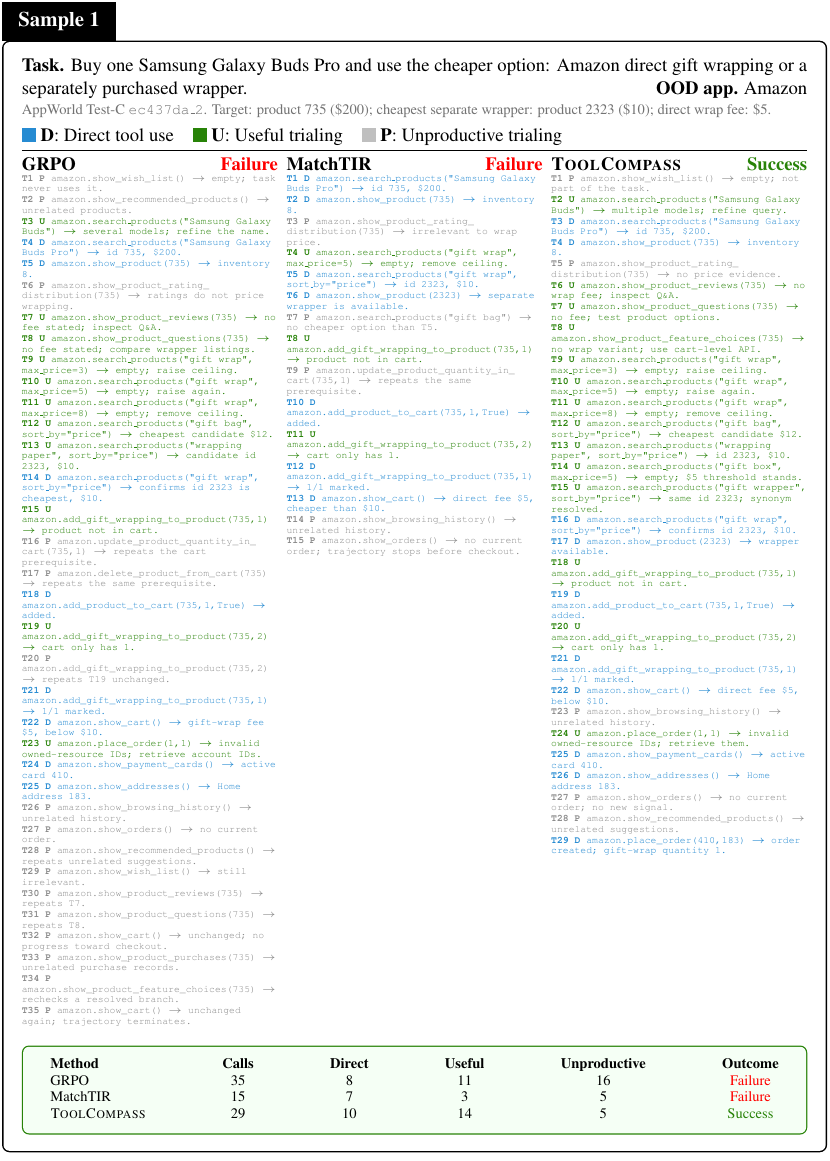}

\caption{ \textbf{Sample 1 (Amazon: compare gift-wrapping options and check out).} \method turns each error into a corrective next step and keeps its trials on checkout-relevant APIs, completing the order in 29 calls with only 5 unproductive ones. GRPO finds the same evidence but spends 16 calls on unrelated or repeated queries and exhausts the interaction budget before checkout. MatchTIR stops after 15 calls without attempting checkout.}
\label{fig:qualitative-sample1}

\end{figure}

\begin{figure}[t]
\centering
\includegraphics[width=\linewidth]{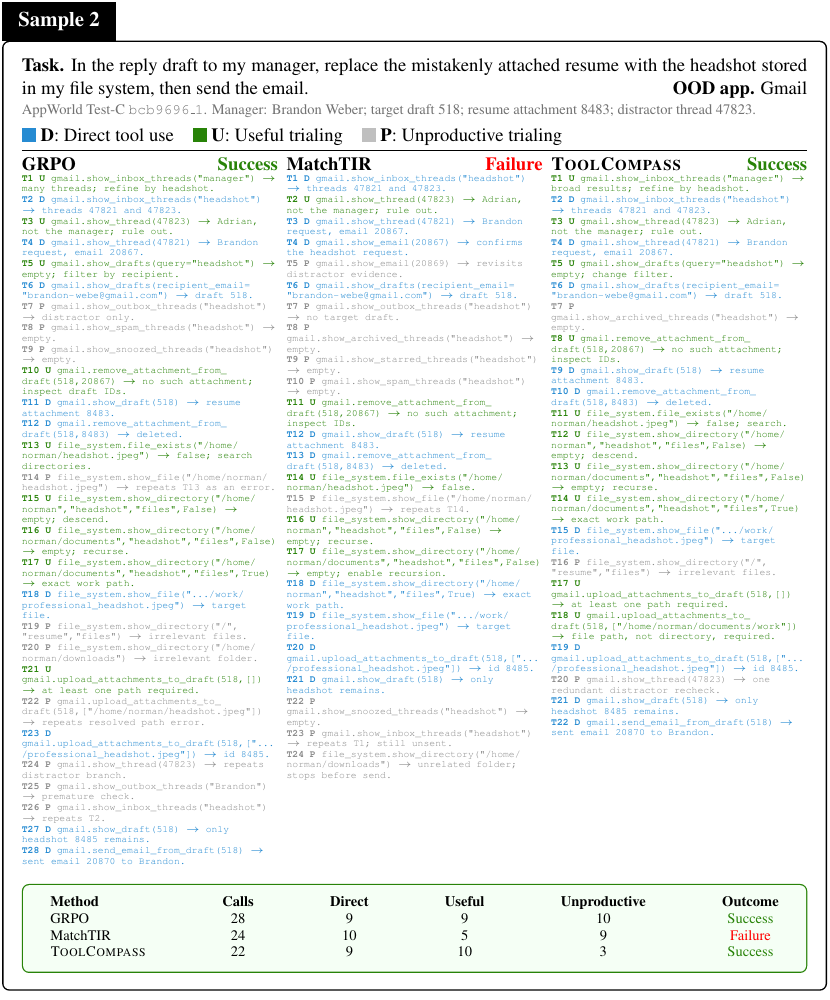}

\caption{ \textbf{Sample 2 (Gmail: replace a draft attachment and send).} All three methods locate the correct thread and draft, but they spend the remaining budget differently. \method resolves the path and upload-format errors with targeted follow-ups and finishes in 22 calls with only 3 unproductive ones. GRPO also succeeds but spends 10 of its 28 calls on unrelated mailbox checks and repeats. MatchTIR ends after 24 calls without sending the email.}
\label{fig:qualitative-sample2}

\end{figure}

We present two AppWorld Test-C case studies to show how \method guides tool trialing on tasks that require the unseen applications. For each task, we show the complete trajectories of GRPO, MatchTIR, and \method, and every call is labeled as direct tool use, useful trialing, or unproductive trialing following the protocol in Appendix~\ref{app:tool-call-categorization}. The task instructions, resource identifiers, and environment observations come from the executable benchmark instances; authentication and API-documentation calls are omitted. Sample~1 (Figure~\ref{fig:qualitative-sample1}) requires comparing two gift-wrapping options and completing a checkout on Amazon. Sample~2 (Figure~\ref{fig:qualitative-sample2}) requires replacing an attachment in a Gmail draft with a file from the file system.

The two cases show the same pattern from different sides. \method explores when exploration is needed: its trials stay on functionally related APIs, and each error is followed by a corrective call. GRPO explores without guidance, which wastes the budget in Sample~2 and causes a failure in Sample~1. MatchTIR suppresses exploration and terminates before the required final actions in both cases. This matches the aggregate behavior in Figure~\ref{fig:tool-trialing}: \method keeps the useful trials while cutting the unproductive ones.

\subsection{AppWorld SGC Results}
\label{app:sgc}

In addition to the task success rate, we report Scenario Goal Completion (SGC), a stricter metric that credits a scenario only when all of its task variants are completed successfully. The results are shown in Table~\ref{tab:appworld-sgc}.

\begin{table}[t]
    \centering
    \caption{AppWorld SGC results (\%) using Qwen3.5-4B.}
    \label{tab:appworld-sgc}
\normalsize
    \setlength{\tabcolsep}{10pt}
    \renewcommand{\arraystretch}{1.04}
    \begin{tabular}{@{}lccc@{}}
        \toprule
        Method & ID & OOD & Total \\
        \midrule
        LOOP & 39.88 & 24.22 & 28.72 \\
        SEAL & 41.07 & 26.38 & 30.60 \\
        \textbf{\method} & \textbf{48.21} & \textbf{29.50} & \textbf{34.87} \\
        \bottomrule
    \end{tabular}
    
\end{table}

As shown in Table~\ref{tab:appworld-sgc}, \method achieves the best SGC on all splits. In particular, it improves OOD SGC by 5.28 points over LOOP and by 3.12 points over SEAL. These results indicate that the gains of \method extend beyond individual tasks and remain consistent across the task variants of the same scenario.

\subsection{FTRL Solve-P, Solve-R, and Solve-F1 Results}
\label{app:ftrl-pr}

We compare \method with LOOP and SEAL on FTRL using Solve-P and Solve-R to distinguish tool-invocation precision from task-completion recall. Table~\ref{tab:ftrl-pr} reports Solve-P, Solve-R, and Solve-F1 on the ID, OOD, and entire test sets using Qwen3.5-4B.

\begin{table}[t]
    
    \centering
    \caption{FTRL results (\%) using Qwen3.5-4B. Each cell reports Solve-P / Solve-R / Solve-F1.}
    \label{tab:ftrl-pr}
\normalsize
    \setlength{\tabcolsep}{8pt}
    \renewcommand{\arraystretch}{1.04}
    \begin{tabular}{@{}lccc@{}}
        \toprule
        Method & ID & OOD & Total \\
        \midrule
        LOOP & 37.80 / 57.60 / 41.83 & 51.90 / 76.40 / 57.72 & 40.06 / 60.61 / 44.37 \\
        SEAL & 33.40 / 56.70 / 38.02 & 53.70 / 73.80 / 57.94 & 36.65 / 59.44 / 41.21 \\
        \textbf{\method} & \textbf{43.20 / 63.70 / 46.97} & \textbf{59.50 / 80.20 / 64.01} & \textbf{45.81 / 66.34 / 49.70} \\
        \bottomrule
    \end{tabular}
    
\end{table}

\method achieves the best results on all three metrics across the ID, OOD, and total test sets. The gains cover both tool-invocation precision and task-completion recall.

\clearpage
\begin{figure}[t]
\begin{promptbox}[unbreakable]{Tool-Call Categorization Prompt}
You label the current call using only the task, tool specifications, prior
history, current call, and its immediate feedback. Choose one label:

- DIRECT_TOOL_USE: Directly performs a required operation or obtains
  information needed to complete the task.
- USEFUL_TRIALING: A plausible exploration that provides new, task-relevant
  information about tool capabilities, arguments, or execution conditions.
- UNPRODUCTIVE_TRIALING: An unrelated or redundant call, or an exploration
  that provides no new task-relevant information.

Apply DIRECT_TOOL_USE first. Assess novelty against the tool specifications
and prior history. Errors can be useful if they reveal new relevant information.
Do not use later actions or the final task outcome.

Input:
task: {task_instruction}
tools: {tool_specifications}
history: {prior_interaction_history}
call: {current_call}
feedback: {current_feedback}

Return strict JSON only:
{"turn": <turn_id>, "label": "<ONE_LABEL>", "reason": "<one sentence>"}
\end{promptbox}

\promptcaption{Prompt used to categorize tool calls with GPT-5.6-sol.}{fig:tool-call-prompt}
\end{figure}

\subsection{Tool-Call Categorization}
\label{app:tool-call-categorization}

For the tool-use behavior analysis in Figure~\ref{fig:tool-trialing} and Appendix~\ref{app:qualitative}, GPT-5.6-sol serves as the LLM-as-a-judge~\citep{zheng2023judging} and assigns each tool call to one of three categories: \emph{direct tool use}, which performs a required operation or obtains information needed to complete the task; \emph{useful trialing}, a plausible exploratory call that provides new, task-relevant information about tool capabilities, argument requirements, or execution conditions; and \emph{unproductive trialing}, which is unrelated, redundant, or provides no new task-relevant information. The judge receives the task instruction, tool specifications, prior interaction history, current call, and its immediate feedback. The full prompt is shown in Figure~\ref{fig:tool-call-prompt}.

Information novelty is assessed against the supplied tool specifications and prior interaction history. Subsequent actions and final task outcomes are excluded from the judgment.

\par

\par

To assess annotation quality, three annotators independently verify a random sample of 200 tool calls with the same information provided to the judge, and we compute Fleiss's kappa over the three sets of labels. The kappa score is 0.73, with a raw agreement rate of 82.00\%; the remaining disagreements are resolved by discussion. Moreover, the LLM annotations agree with the final reviewed labels on 86.00\% of the calls, supporting their use for the full behavior analysis.

\subsection{Representation-Space Visualization Details}
\label{app:vis_details}

For the visualization in Figure~\ref{fig:vmf_sphere}, we select four function classes and extract their raw layer-\(l=8\) backbone hidden states, without applying the projection head. These states are normalized and mapped onto the three-dimensional unit sphere for rendering. The two panels share the same viewpoint and a jointly normalized color scale, and the surface shows the spherical kernel density of the displayed class.

\end{document}